\documentclass[11pt]{article}

\usepackage[final]{acl}
\usepackage{times}
\usepackage{latexsym}
\usepackage[T1]{fontenc}
\usepackage[utf8]{inputenc}
\usepackage{microtype}
\usepackage{inconsolata}

\usepackage{graphicx}
\usepackage{booktabs}
\usepackage{multirow}
\usepackage{amsmath}
\usepackage{amssymb}
\usepackage{tikz}
\usepackage{pgfplots}
\pgfplotsset{compat=1.18}
\usetikzlibrary{patterns}
\usepgfplotslibrary{groupplots}
\usepackage{colortbl}
\usepackage{xcolor}
\usepackage{oplotsymbl}

\definecolor{palRed}{HTML}{B9181A}
\definecolor{palOrangeRed}{HTML}{EE4431}
\definecolor{palOrange}{HTML}{FC8002}
\definecolor{palLightOrange}{HTML}{FABB6E}
\definecolor{palGreen}{HTML}{369F2D}
\definecolor{palLightGreen}{HTML}{ADDB88}
\definecolor{palSalmon}{HTML}{FAC7B3}
\definecolor{palDarkPurple}{HTML}{614099}
\definecolor{palMidPurple}{HTML}{8481BA}
\definecolor{palLightPurple}{HTML}{B4B4D5}
\definecolor{palDarkBlue}{HTML}{1663A9}
\definecolor{palBlue}{HTML}{4995C6}
\definecolor{palLightBlue}{HTML}{92C2DD}
\definecolor{palVeryLightBlue}{HTML}{CEDFEF}

\title{Beyond Controllability: Speaker Diversity in Instruction-Guided Text-to-Speech}

\author{
  Yi-Cheng Lin\textsuperscript{1},
  Jia-Kai Dong\textsuperscript{1},
  Kuan-Yu Chen\textsuperscript{1},
  Szu-Chi Chen\textsuperscript{1},\\
  \textbf{
  Jian-Jiun Ding\textsuperscript{1},
  Sung-Feng Huang\textsuperscript{2},
  Yusuke Hirota\textsuperscript{2},
  Hung-yi Lee\textsuperscript{1,3}}
  \\
  \textsuperscript{1}National Taiwan University \quad
  \textsuperscript{2}NVIDIA \\
  \textsuperscript{3}Artificial Intelligence Center of Research Excellence, National Taiwan University
  \\
  \texttt{\{f12942075, hungyilee\}@ntu.edu.tw}
}

\begin{document}
\maketitle

\begin{abstract}
Instruction-guided text-to-speech (ITTS) lets users specify voice properties in natural language.
Applications from conversational AI character voices to synthetic training data for downstream speech models depend on the system producing many distinct voices when the same broad category is requested repeatedly.
No prior work audits whether ITTS outputs are in fact diverse, either compared to real speech or across attribute values.
We evaluate seven open-source ITTS systems using speaker embeddings whose representational capacity we independently validate.
Every system produces speaker distributions markedly narrower than real speech.
Furthermore, female prompts produce less diverse voice pools than male prompts across most systems.
In contrast, accent and emotion prompts barely alter speaker diversity in most systems, but reveal a strong controllability bias toward North American English and neutral speech.
These results show that diversity and controllability capture complementary weaknesses, neither of which alone is sufficient to evaluate ITTS.
Our code and data are available at \url{https://github.com/dlion168/ITTS-Speaker-Diversity}.
\end{abstract}

\section{Introduction}
\label{sec:intro}

\begin{figure*}
    \centering
    \includegraphics[width=0.90\textwidth]{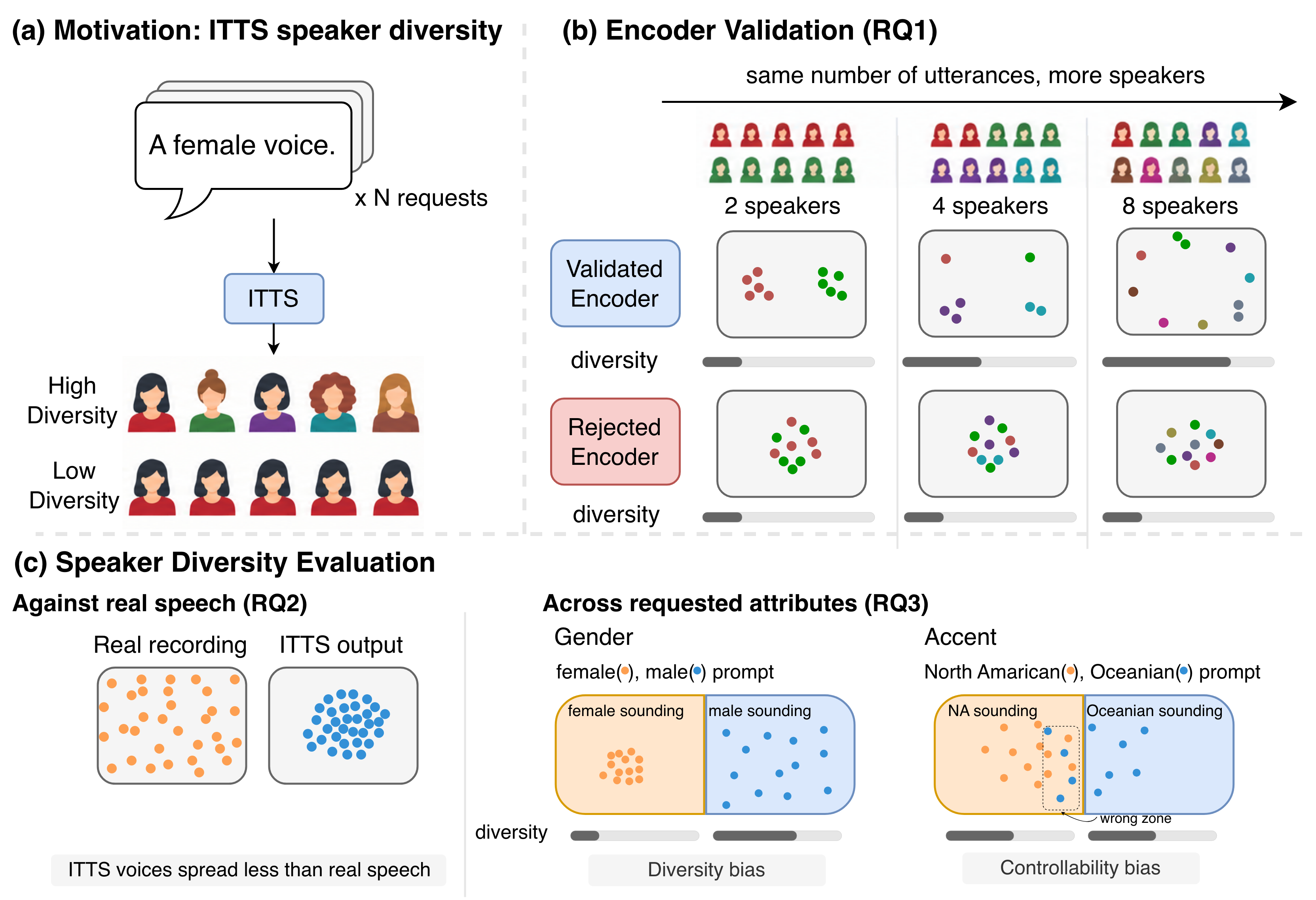}
    \caption{Overview. (a) A broad description should produce many distinct voices. (b) We keep only encoders whose diversity scores track speaker count (RQ1). (c) ITTS voices are less diverse than real speech (RQ2), and diversity and controllability vary across voice attributes (RQ3).}
    \label{fig:overview}
\end{figure*}

Instruction-guided text-to-speech (ITTS) has attracted growing attention in recent years \citep{xie2024controllable, wu2026mint, evoc2025}.
In contrast to the traditional paradigms of selecting a fixed speaker embedding \citep{wang2018styletokens} or cloning a voice from reference audio \citep{wang2023valle, chen2024f5tts}, ITTS allows users to specify voice properties such as gender, age, accent, and speaking style directly in natural language.
Downstream applications critically depend on output speaker diversity, as conversational AI requires synthesizing distinct personas without manual voice engineering, while training downstream speech models requires generating synthetic corpora with broad speaker coverage \citep{wagner2025personalized, schuhmann2025emonetvoice}.
Because writing a unique description for every voice is impractical at scale, corpus synthesis relies on the system to produce different voices when the same broad description, such as ``a female voice’’, is requested repeatedly (Figure~\ref{fig:overview}a).

However, existing evaluations of ITTS diversity leave two gaps.
First, prior studies do not compare speaker diversity to a real-speech baseline.
Second, prior studies overlook how diversity varies across values of voice attributes such as gender, age, or accent.
If a model generates only a few repetitive voices for certain groups like children or minority accents, synthetic datasets created from it will carry that same bias and degrade the performance of downstream speech models \citep{lin2026fair}.

We ask three questions, summarized in Figure~\ref{fig:overview}b,c.
\textbf{RQ1}: Which speaker encoders are suitable for measuring ITTS speaker diversity? 
\textbf{RQ2}: Do ITTS systems generate voices as diverse as those in real speech? 
\textbf{RQ3}: Is ITTS speaker diversity similar across attribute values, such as female and male? 
For RQ1, we test whether each encoder’s diversity scores clearly increase with the number of distinct speakers in a set.
For RQ2, we use the validated encoders to compare ITTS outputs with real recordings of the same texts.
For RQ3, we compare speaker diversity and controllability across prompts that request different genders, ages, accents, and emotions.

Our work yields three primary contributions:
\begin{itemize}
\itemsep0pt
\item \textbf{Methodological validation (RQ1):} We show that several widely used speaker encoders are unsuitable for measuring speaker diversity, because their diversity scores barely differ between sets with few and many speakers.
\item \textbf{Quantification of the reality gap (RQ2):} We find that every evaluated open-source ITTS system produces voices less diverse than real human voices.
\item \textbf{Detection of biases (RQ3):} We show that speaker diversity and controllability capture complementary biases across voice attribute prompts.
Most systems yield lower speaker diversity for female prompts despite accurate gender control, whereas accent and emotion prompts preserve diversity but exhibit controllability biases toward North American English and neutral speech.
\end{itemize}

\section{Related Work}
\label{sec:related}

\paragraph{Instruction-guided TTS.} 
ITTS systems control speech through natural-language instructions but differ in how they determine speaker identity. 
One line of work relies on reference audio or preset voices to fix the speaker, using text instructions to control speaking style, prosody, or emotion \citep{ji2025controlspeech, du2024cosyvoice1, du2024cosyvoice, du2025cosyvoice3, zhou2025indextts2}.
In contrast, description-only generation synthesizes the speaker entirely from natural language prompts without external acoustic references, an approach supported by an expanding set of models \citep{guo2023prompttts, zhang2023promptspeaker, vyas2023audiobox, lyth2024parler, zhou2024voxinstruct, capspeech2025, li2025flespeech, minimax2025speech, openmoss2026voicegen, qwen3tts2026, voxcpm2_2026}.
Several of these models also accept a reference recording \citep{vyas2023audiobox, li2025flespeech, minimax2025speech, zhou2024voxinstruct, voxcpm2_2026}, but we evaluate every system in description-only mode.
Because a text description leaves acoustic details underspecified \citep{leng2024prompttts2}, the model must determine the remaining vocal traits.

\paragraph{Diversity evaluation for generative speech.} Prior evaluations of speech diversity focus either on training corpora, distributional realism, or global demographic spread in zero-shot systems, leaving fine-grained, attribute-conditional speaker diversity in ITTS largely unexamined. 
GLOBE \citep{wang2024globe} applies the Vendi Score to TitaNet speaker embeddings, but on TTS \emph{training corpora} rather than TTS model \emph{outputs}.
TTSDS \citep{minixhofer2024ttsds,minixhofer2025ttsds2} measures realism via distributional distance to a real reference. It doesn't report within-set diversity or analyze how this diversity varies across requested voice attribute values.
MAD-Speech \citep{futeral2025mad} evaluates the global diversity of conventional TTS systems whose voices are fixed, reference-based, or randomly sampled.
In contrast, we target intra-prompt diversity in ITTS, whose voice can be steered by natural language instructions. 
MINT-Bench \citep{wu2026mint}, a benchmark for instruction-following TTS, includes a Timbre Diversity Score. 
For a given instruction and text, it generates several outputs and measures how dissimilar the WavLM-Large speaker embeddings of those that follow the instruction are. 
The benchmark summarizes this as a single score per system and language, so it does not show how diversity differs across attribute values or how it compares with real speech.

\paragraph{Bias evaluation in TTS.}
Prior work on TTS bias has focused on three main dimensions.
The first is unwanted correlation between unrelated prompt attributes and voice, such as occupation-to-gender stereotype leakage \citep{kuan2025gender,bindingeffect2026}.
The second is unequal output quality across demographic groups, such as perceptual quality gaps across accents \citep{michel2025accent}.
The third is demographic controllability, where E-VOC \citep{evoc2025} shows that systems often fail to follow child and elderly prompts.
Our work investigates controllability alongside a fourth dimension, disparities in speaker diversity across requested voice attribute values.
We show that evaluating both controllability and diversity is essential, as they capture complementary biases that neither metric reveals alone.





\section{Method}
\label{sec:method}
 
We define speaker diversity as the variation in speaker identity among the outputs a system generates under one instruction condition. 
Under a generic instruction that specifies no speaker attribute (``Speak like a person---natural, neutral tone, moderate pace, conversational and clear delivery.’’), this gives the overall speaker diversity of a system (\S\ref{sec:overall}). 
Under instructions that request an attribute value, such as a female voice, it gives the \emph{attribute-conditional speaker diversity} for that value (\S\ref{sec:conditional}).
We quantify this variation within speaker-verification embedding spaces, which are explicitly optimized to capture speaker-discriminative characteristics.
 
\subsection{Speaker Diversity Metrics}
\label{sec:metric}
 
Given $N$ L2-normalized speaker embeddings $\{x_1, \dots, x_N\}$, let $K$ be the $N \times N$ cosine-similarity matrix with $K_{ij} = x_i^\top x_j$.
We summarize this matrix with two diversity metrics.
The first is the mean pairwise cosine distance (MPCD),
\begin{equation}
    \mathrm{MPCD}(X) = \tfrac{1}{N(N-1)} \sum_{i \neq j} (1 - K_{ij}).
\end{equation}
The second is the Vendi Score \citep{friedman2023vendi},
\begin{equation}
\mathrm{VS}(X) = \exp\!\left( -\sum_{\lambda_l > 0} \lambda_l \log \lambda_l \right),
\end{equation}
where $\lambda_1, \dots, \lambda_N$ are the eigenvalues of $K/N$ (satisfying $\sum_l \lambda_l = 1$).
Vendi is the exponential of the entropy of these eigenvalues, reflecting the effective number of orthogonal modes in the speaker-embedding space.
 
Both pairwise averages and the Vendi Score are widely used to measure the diversity of generated images \citep{sadat2024cads, askarihemmat2024improving} and text \citep{tevet2021evaluating, jalali2024cvendi}.
MPCD averages over all pairs, so a set concentrated on a few well-separated voices still scores high.
Vendi depends on the full eigenvalue spectrum and drops toward the number of such voices, so we treat it as the primary measure and report MPCD as a complementary check.
 
Prior work also evaluates generated audio by its distributional distance to real recordings.
Because a large distance can reflect either limited voice diversity or poor acoustic realism, these metrics do not measure diversity alone, so we defer them to Appendix~\ref{app:mad}.

\begin{figure*}[t]
\centering
\begin{tikzpicture}
\pgfplotsset{
  validation-panel/.style={
    width=7.6cm, height=4.4cm,
    xmode=log,
    xmin=5.5, xmax=260,
    xtick={6,10,20,40,80,240},
    xticklabels={6,10,20,40,80,240},
    tick label style={font=\scriptsize},
    label style={font=\footnotesize},
    title style={font=\footnotesize},
    xlabel={ground-truth $k$ (distinct speakers)},
    grid=both, grid style={draw=gray!15},
    cycle list={
      {color=palRed,         mark=*,        mark size=1.2pt, thick},
      {color=palOrangeRed,   mark=square*,  mark size=1.2pt, thick},
      {color=palOrange,      mark=triangle*,mark size=1.4pt, thick},
      {color=palLightOrange, mark=diamond*, mark size=1.4pt, thick},
      {color=palGreen,       mark=*,        mark size=1.2pt, thick},
      {color=palLightGreen,  mark=square*,  mark size=1.2pt, thick},
      {color=palSalmon,      mark=triangle*,mark size=1.4pt, thick},
      {color=palDarkPurple,  mark=diamond*, mark size=1.4pt, thick},
      {color=palDarkBlue,    mark=x,        mark size=2pt,   thick, dashed},
      {color=palBlue,        mark=+,        mark size=2pt,   thick, dashed},
      {color=palLightBlue,   mark=o,        mark size=1.4pt, thick, dashed},
    },
  }
}
\begin{groupplot}[
  group style={group size=2 by 1, horizontal sep=1.4cm},
  validation-panel,
]
\nextgroupplot[title={MPCD ($\uparrow$ with $k$)}, ylabel={MPCD}, ymin=0, ymax=1]
\addplot coordinates {(6,0.825) (8,0.8606) (10,0.8775) (12,0.8904) (15,0.9053) (20,0.9184) (24,0.9264) (30,0.9316) (40,0.9397) (48,0.9413) (60,0.9464) (80,0.9488) (120,0.9521) (240,0.9556)};
\addplot coordinates {(6,0.8053) (8,0.8351) (10,0.8521) (12,0.8657) (15,0.8757) (20,0.8869) (24,0.8942) (30,0.8992) (40,0.9054) (48,0.9082) (60,0.9114) (80,0.9144) (120,0.9172) (240,0.9199)};
\addplot coordinates {(6,0.7898) (8,0.8197) (10,0.8345) (12,0.8485) (15,0.8584) (20,0.8692) (24,0.8758) (30,0.8814) (40,0.8877) (48,0.8907) (60,0.8939) (80,0.8961) (120,0.8989) (240,0.902)};
\addplot coordinates {(6,0.5131) (8,0.5304) (10,0.5382) (12,0.5459) (15,0.5518) (20,0.5596) (24,0.5654) (30,0.568) (40,0.5728) (48,0.574) (60,0.5783) (80,0.5793) (120,0.581) (240,0.5833)};
\addplot coordinates {(6,0.7832) (8,0.8131) (10,0.8284) (12,0.8431) (15,0.8542) (20,0.8674) (24,0.8734) (30,0.8796) (40,0.8858) (48,0.889) (60,0.8923) (80,0.895) (120,0.8975) (240,0.9007)};
\addplot coordinates {(6,0.5161) (8,0.5364) (10,0.5432) (12,0.5516) (15,0.5578) (20,0.5648) (24,0.5706) (30,0.5715) (40,0.5781) (48,0.5798) (60,0.5833) (80,0.5828) (120,0.5855) (240,0.5882)};
\addplot coordinates {(6,0.7657) (8,0.7962) (10,0.8141) (12,0.8267) (15,0.8378) (20,0.8529) (24,0.859) (30,0.864) (40,0.8704) (48,0.8735) (60,0.8773) (80,0.8802) (120,0.8826) (240,0.8861)};
\addplot coordinates {(6,0.5166) (8,0.534) (10,0.5403) (12,0.5493) (15,0.5553) (20,0.5638) (24,0.5691) (30,0.5706) (40,0.5767) (48,0.5776) (60,0.5815) (80,0.583) (120,0.5849) (240,0.5868)};
\addplot coordinates {(6,0.07908) (8,0.08205) (10,0.08385) (12,0.08489) (15,0.0857) (20,0.08655) (24,0.08779) (30,0.08777) (40,0.08889) (48,0.08894) (60,0.08937) (80,0.08978) (120,0.08995) (240,0.09032)};
\addplot coordinates {(6,0.2649) (8,0.2805) (10,0.2823) (12,0.2863) (15,0.2927) (20,0.2982) (24,0.3005) (30,0.3008) (40,0.3027) (48,0.3068) (60,0.3054) (80,0.3082) (120,0.3088) (240,0.3104)};
\addplot coordinates {(6,0.2609) (8,0.2744) (10,0.2773) (12,0.2804) (15,0.287) (20,0.2931) (24,0.2962) (30,0.2965) (40,0.2979) (48,0.3017) (60,0.3008) (80,0.3042) (120,0.304) (240,0.3056)};
\nextgroupplot[title={Vendi Score ($\uparrow$ with $k$)}, ylabel={Vendi}, legend to name=validationlegend,
  legend columns=11,
  legend style={font=\tiny, draw=none, column sep=3pt},
]
\addplot coordinates {(6,12.01) (8,15.24) (10,18.19) (12,20.64) (15,24.38) (20,29.47) (24,33.29) (30,37.95) (40,44.71) (48,48.92) (60,54.28) (80,60.9) (120,69.25) (240,80.5)}; \addlegendentry{TIT}
\addplot coordinates {(6,15.81) (8,19.62) (10,23.13) (12,26) (15,30.26) (20,35.72) (24,40.07) (30,45.07) (40,52.23) (48,56.89) (60,62.56) (80,69.66) (120,78.87) (240,91.44)}; \addlegendentry{SBRN}
\addplot coordinates {(6,15.4) (8,19.13) (10,22.26) (12,25.2) (15,29.09) (20,34.16) (24,38.12) (30,42.77) (40,49.3) (48,53.49) (60,58.62) (80,64.7) (120,72.65) (240,83.26)}; \addlegendentry{SBEC}
\addplot coordinates {(6,7.505) (8,8.784) (10,9.769) (12,10.6) (15,11.67) (20,13.03) (24,14.05) (30,15.05) (40,16.42) (48,17.19) (60,18.23) (80,19.18) (120,20.36) (240,21.8)}; \addlegendentry{ER2}
\addplot coordinates {(6,13.09) (8,16.37) (10,19.33) (12,21.89) (15,25.64) (20,30.59) (24,34.41) (30,38.97) (40,45.5) (48,49.85) (60,55.05) (80,61.48) (120,70) (240,81.76)}; \addlegendentry{WS221}
\addplot coordinates {(6,7.766) (8,9.069) (10,10.02) (12,10.85) (15,11.88) (20,13.13) (24,14.07) (30,14.95) (40,16.29) (48,16.98) (60,17.92) (80,18.66) (120,19.74) (240,21.05)}; \addlegendentry{CAM}
\addplot coordinates {(6,11.03) (8,13.9) (10,16.53) (12,18.78) (15,22.02) (20,26.54) (24,29.92) (30,33.9) (40,39.78) (48,43.58) (60,48.16) (80,53.92) (120,61.18) (240,71.07)}; \addlegendentry{WSVB}
\addplot coordinates {(6,7.65) (8,8.939) (10,9.921) (12,10.79) (15,11.88) (20,13.31) (24,14.31) (30,15.28) (40,16.75) (48,17.51) (60,18.55) (80,19.56) (120,20.79) (240,22.24)}; \addlegendentry{ER2v2}
\addplot coordinates {(6,1.609) (8,1.654) (10,1.684) (12,1.706) (15,1.727) (20,1.75) (24,1.772) (30,1.783) (40,1.807) (48,1.814) (60,1.827) (80,1.838) (120,1.847) (240,1.863)}; \addlegendentry{XV}
\addplot coordinates {(6,2.672) (8,2.866) (10,2.952) (12,3.036) (15,3.132) (20,3.247) (24,3.301) (30,3.358) (40,3.417) (48,3.474) (60,3.486) (80,3.529) (120,3.557) (240,3.609)}; \addlegendentry{WLM}
\addplot coordinates {(6,2.493) (8,2.629) (10,2.704) (12,2.762) (15,2.839) (20,2.936) (24,2.98) (30,3.016) (40,3.057) (48,3.102) (60,3.109) (80,3.147) (120,3.16) (240,3.2)}; \addlegendentry{USAT}
\end{groupplot}
\end{tikzpicture}

\vspace{2pt}
\centerline{\ref{validationlegend}}

\caption{\textbf{Encoder calibration with MPCD and Vendi.} Mean MPCD and Vendi over 100 random subsets at each ground-truth speaker count $k$, for all 11 candidate encoders. Solid lines denote the eight encoders meeting calibration criteria and retained for evaluation. Dashed lines indicate the three excluded candidates.}
\label{fig:validation}
\end{figure*}
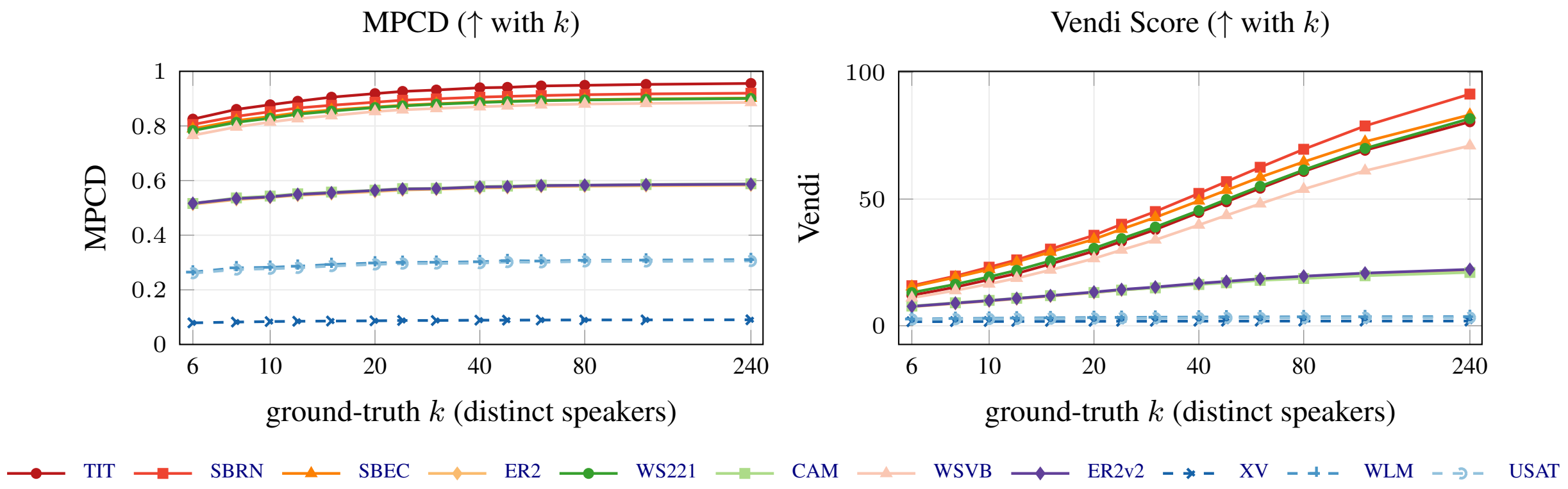
\subsection{Speaker Encoders and Validation}
\label{sec:encoder-val}

Because diversity is measured in an embedding space, encoder selection is critical.
We evaluate \textbf{11 speaker encoders} across three architectural families, each given a short abbreviation used in subsequent tables.
Nine encoders are trained from scratch for speaker verification on labeled speaker data.
These are NeMo-TitaNet (TIT) \citep{koluguri2022titanet}, SpeechBrain-ECAPA (SBEC) and SpeechBrain-ResNet (SBRN) \citep{desplanques2020ecapa}, WeSpeaker-ResNet221 (WS221) and WeSpeaker-SimAM-ResNet34 (WSVB) \citep{wang2023wespeaker}, ModelScope-ERes2Net (ER2), ERes2NetV2 (ER2v2) \cite{chen24l_interspeech}, CAM++ (CAM) \cite{wang23ha_interspeech}, and x-vector (XV) \citep{snyder2018xvector}.
The remaining two start from SSL foundation models pretrained on large-scale unlabeled speech and add a speaker-verification head fine-tuned on top.
These are WavLM-base+SV (WLM) and UniSpeech-SAT-base+SV (USAT) \cite{chen2022unispeech}.

To identify faithful encoders for speaker diversity, we adapt the calibration approach from MAD-Speech \citep{futeral2025mad}.
We construct controlled subsets from LibriSpeech \cite{librispeech} train-clean-100 with varying numbers of unique speakers $k \in \{6, 8, 10, 12, 15, 20, 24, 30, 40, 48, 60,\allowbreak 80,\allowbreak 120,\allowbreak 240\}$.
For each $k$, we build 100 subsets of $N = 240$ utterances by sampling $240/k$ utterances from each of the $k$ speakers.
We compute MPCD and VS on each subset, and plot the resulting calibration curves against ground-truth $k$ in Figure~\ref{fig:validation}.

Across the evaluated range, all eleven encoders produce mean VS curves that increase monotonically with $k$, yielding strong rank correlation (detailed in Appendix~\ref{app:mad}).
However, monotonic scaling alone is insufficient. A reliable encoder must also yield differences large enough to distinguish neighboring speaker counts beyond subset-level sampling noise.
In Figure~\ref{fig:validation}, the encoders clearly divide into two groups.
Eight of the models exhibit substantial dynamic range, expanding their mean VS by a factor of 2.7 to 6.7 as speaker count increases from $k=6$ to $k=240$. Furthermore, across the 100 subsets per $k$, the difference in mean VS between consecutive speaker counts exceeds the subset standard deviation in at least 12 of the 13 adjacent steps ($k_i$ to $k_{i+1}$).
In contrast, XV, WLM, and USAT (dashed lines), remain severely compressed. Across the same range from $k=6$ to $k=240$, their mean VS rises by at most a factor of 1.35 and stays below 4 even at 240 speakers, with the gap between consecutive steps exceeding one standard deviation in at most 1 of the 13 transitions.
Because their scaling under cosine similarity is overshadowed by subset-level sampling noise, we exclude these three checkpoints and conduct all subsequent experiments using the eight retained encoders.

\subsection{ITTS Panel}
\label{sec:panel}

Our evaluation panel comprises seven open-source models spanning four architectural approaches.
Four models are codec-based autoregressive (AR) language models that generate discrete audio tokens sequentially: MOSS-VoiceGenerator \citep{openmoss2026voicegen}, Parler-TTS Mini v1 and Parler-TTS Large v1 \citep{lyth2024parler}, and Qwen3-TTS Voice Design \citep{qwen3tts2026}.
VoxInstruct \citep{zhou2024voxinstruct} combines AR and non-autoregressive (NAR) generation over discrete semantic and acoustic tokens.
CapSpeech-NAR \citep{capspeech2025} generates speech non-autoregressively through flow matching over a Diffusion Transformer (DiT) backbone.
Finally, VoxCPM2 \citep{voxcpm2_2026} generates continuous latent patches sequentially, with each patch produced by a Diffusion Transformer conditioned on previously generated patches.

\subsection{Prompt Design and Sampling}
\label{sec:sampling}

We evaluate speaker diversity across five attribute axes, spanning gender, age, accent, emotion, and an intersectional gender-by-age axis.
Gender covers female and male voices, while emotion includes angry, happy, neutral, and sad states.
Age is partitioned into four groups covering children aged 19 and under, young adults aged 20 to 39, middle-aged adults aged 40 to 59, and elderly aged 60 and above \cite{hales2017prevalence}.
The accent axis groups regional varieties rather than national borders, including North American, British, Oceanian, South Asian, South African, and Germanic varieties to probe coverage across historically underrepresented groups.
Because gender and age exhibit the most pronounced diversity disparities in our study, we combine them into an intersectional axis covering all eight female and male pairings across the four age bins to analyze compound biases.

We design five prompt templates for each single-attribute value, and ten for each intersectional gender and age combination.
All prompt templates, sample realizations, and our construction rationale are detailed in Appendix~\ref{app:prompts}. Instructions are passed through each system's official prompt interface, and sampling parameters follow each system's default settings unless otherwise specified.

We draw text and audio pairs from the English subset of the Emilia corpus \citep{emilia}, using the text prompts for synthesis and the original recordings as our real-speech reference.
Because Emilia collects real-world audio across podcasts, debates, and talk shows, it spans a far wider variety of speakers than traditional studio datasets.
This broad coverage establishes a challenging baseline to test whether modern ITTS systems produce less diverse voices than real speech.
For the overall diversity evaluation, we sample 30{,}000 sentences and partition them into three disjoint sets of 10{,}000. 
We compute Vendi separately on each set and report the mean and standard deviation over the three sets, applying the same protocol to the paired real recordings.
For attribute conditioning (\S\ref{sec:conditional}), we sample a set of 3{,}000 texts. These texts are shared across all attribute values to hold linguistic content constant.

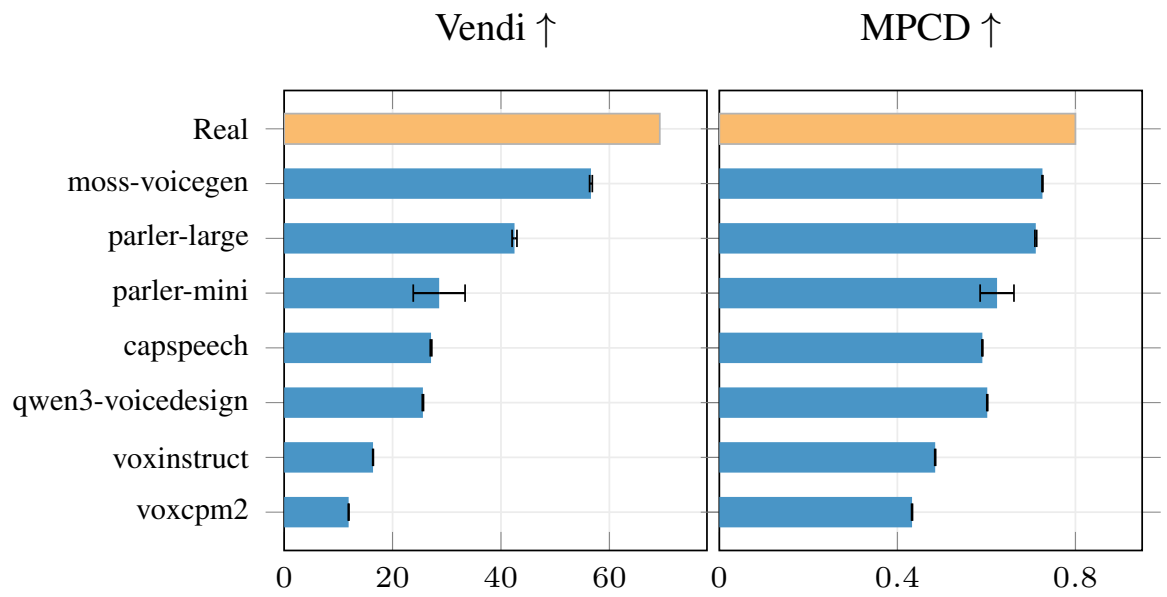
\begin{figure}[t]
\centering
\resizebox{\linewidth}{!}{%
\begin{tikzpicture}
\pgfplotsset{
  bar-panel/.style={
    xbar,
    bar width=7pt,
    bar shift=0pt,
    width=5cm,
    height=5.3cm,
    enlarge y limits=0.1,
    tick label style={font=\scriptsize},
    title style={font=\footnotesize},
    ytick={voxcpm2,voxinstruct,qwen3-voicedesign,capspeech,parler-mini,parler-large,moss-voicegen,Real},
    yticklabels={voxcpm2,voxinstruct,qwen3-voicedesign,capspeech,parler-mini,parler-large,moss-voicegen,Real},
    symbolic y coords={voxcpm2,voxinstruct,qwen3-voicedesign,capspeech,parler-mini,parler-large,moss-voicegen,Real},
    grid=major, grid style={draw=gray!15},
  }
}
\begin{groupplot}[
  group style={group size=2 by 1, horizontal sep=0.1cm},
  bar-panel,
]
\nextgroupplot[title={Vendi $\uparrow$}, xmin=0, xmax=78, xtick={0,20,40,60}]
\addplot[fill=palLightOrange, draw=black!30, bar shift=0pt]
  coordinates {(69.3,Real)};
\addplot[fill=palBlue, draw=none, bar shift=0pt, error bars/.cd, x dir=both, x explicit, error bar style={black, line width=0.5pt}] coordinates {
  (56.6,moss-voicegen) +- (0.25,0)
  (42.5,parler-large) +- (0.45,0)
  (28.6,parler-mini) +- (4.78,0)
  (27.1,capspeech) +- (0.14,0)
  (25.6,qwen3-voicedesign) +- (0.11,0)
  (16.4,voxinstruct) +- (0.06,0)
  (11.9,voxcpm2) +- (0.06,0)
};
\nextgroupplot[title={MPCD $\uparrow$}, xmin=0, xmax=0.95, xtick={0,0.4,0.8}, yticklabels={,,,,,,,,}]
\addplot[fill=palLightOrange, draw=black!30, bar shift=0pt]
  coordinates {(0.800,Real)};
\addplot[fill=palBlue, draw=none, bar shift=0pt, error bars/.cd, x dir=both, x explicit, error bar style={black, line width=0.5pt}] coordinates {
  (0.726,moss-voicegen) +- (0.001,0)
  (0.711,parler-large) +- (0.002,0)
  (0.624,parler-mini) +- (0.038,0)
  (0.591,capspeech) +- (0.001,0)
  (0.602,qwen3-voicedesign) +- (0.001,0)
  (0.485,voxinstruct) +- (0.001,0)
  (0.433,voxcpm2) +- (0.001,0)
};
\end{groupplot}
\end{tikzpicture}}
\caption{\textbf{Overall speaker diversity under the neutral prompt.} ITTS systems (blue) are compared against the Real reference (orange). Error bars are batch-to-batch standard deviations across the three generation batches.}
\label{fig:overall}
\end{figure}

\subsection{Controllability Measurement}
\label{sec:controllability}
Following prior work \citep{capspeech2025, zhang2023promptspeaker, ren2024emocodec}, we measure controllability by applying acoustic attribute classifiers to the generated speech.
For each requested value, controllability is quantified as per-class recall, defined as the proportion of generated utterances correctly identified by the classifier.
The chance baseline is $1/K$ for an attribute with $K$ classes, yielding $1/2$ for gender, $1/4$ for age and emotion, and $1/6$ for accent.
Appendix~\ref{app:classifiers} details the classifier architectures and validation setups for each evaluated attribute.

\section{Overall Speaker Diversity}
\label{sec:overall}
\subsection{Comparison with Real Speech}
\label{sec:overall-narrower}

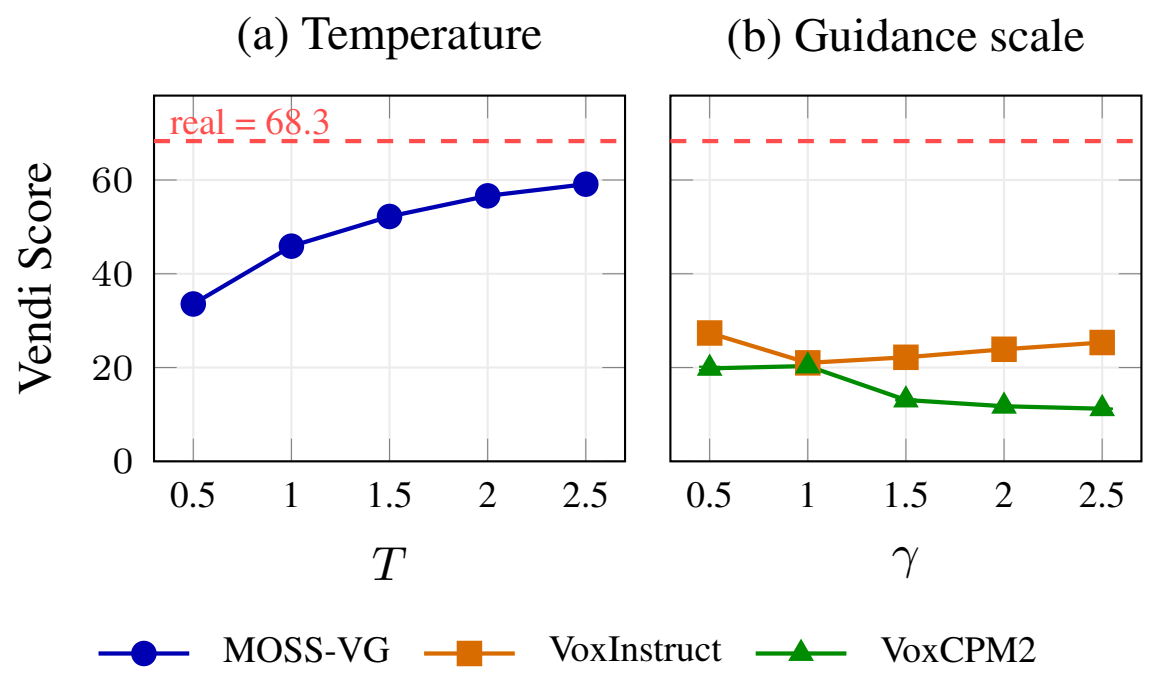
\begin{figure}[t]
\centering
\begin{tikzpicture}
\begin{groupplot}[
  group style={group size=2 by 1, horizontal sep=0.3cm, y descriptions at=edge left},
  width=0.61\linewidth, height=4.0cm,
  ymin=0, ymax=78, ytick={0,20,40,60},
  xmin=0.3, xmax=2.7, xtick={0.5,1.0,1.5,2.0,2.5}, xticklabels={0.5,1,1.5,2,2.5},
  tick label style={font=\scriptsize},
  label style={font=\footnotesize},
  title style={font=\footnotesize, yshift=-3pt},
  grid=major, grid style={draw=gray!15},
]
\nextgroupplot[title={(a) Temperature}, xlabel={$T$}, ylabel={Vendi Score}]
\addplot[mark=*, thick, color=blue!70!black, error bars/.cd, y dir=both, y explicit] coordinates {
  (0.5, 33.56) +- (0, 0.16)
  (1.0, 45.89) +- (0, 0.30)
  (1.5, 52.22) +- (0, 0.44)
  (2.0, 56.62) +- (0, 0.56)
  (2.5, 59.11) +- (0, 0.26)
};
\draw[dashed, thick, red!70] (axis cs:0.3,68.3) -- (axis cs:2.7,68.3);
\node[font=\scriptsize, red!70, anchor=south west, inner sep=1pt] at (axis cs:0.35,68.3) {real = 68.3};
\nextgroupplot[title={(b) Guidance scale}, xlabel={$\gamma$},
  legend to name=sweeplegend, legend columns=3,
  legend style={font=\scriptsize, draw=none, column sep=4pt, text=black}]
\addlegendimage{mark=*, thick, color=blue!70!black}
\addlegendentry{MOSS-VG}
\addplot[mark=square*, thick, color=orange!85!black, error bars/.cd, y dir=both, y explicit] coordinates {
  (0.5, 27.37) +- (0, 0.18)
  (1.0, 21.00) +- (0, 0.09)
  (1.5, 22.19) +- (0, 0.05)
  (2.0, 23.90) +- (0, 0.21)
  (2.5, 25.34) +- (0, 0.21)
};
\addlegendentry{VoxInstruct}
\addplot[mark=triangle*, thick, color=green!55!black, error bars/.cd, y dir=both, y explicit] coordinates {
  (0.5, 19.83) +- (0, 0.35)
  (1.0, 20.33) +- (0, 0.11)
  (1.5, 13.08) +- (0, 0.12)
  (2.0, 11.75) +- (0, 0.03)
  (2.5, 11.22) +- (0, 0.07)
};
\addlegendentry{VoxCPM2}
\draw[dashed, thick, red!70] (axis cs:0.3,68.3) -- (axis cs:2.7,68.3);
\end{groupplot}
\end{tikzpicture}
 
\vspace{2pt}
\centerline{\ref{sweeplegend}}
 
\caption{\textbf{Speaker diversity under different sampling parameters.} Vendi for the three systems that expose decoding controls, under (a) sampling temperature $T$ for MOSS-VoiceGenerator and (b) classifier-free guidance scale $\gamma$ for VoxInstruct and VoxCPM2. Both panels share the same y-axis. Each point is the mean over the eight validated encoders and three seeds. The dashed horizontal line marks the real-speech baseline.}
\label{fig:paramsweep}
\end{figure}
 
As shown in Figure~\ref{fig:overall}, every system exhibits lower diversity than the real-speech reference across both metrics, a trend that holds consistently across each of the eight individual encoders rather than merely on average. Furthermore, the two metrics largely align on the relative model rankings. Among the evaluated systems, MOSS-VoiceGenerator comes closest to real speech, whereas VoxCPM2 suffers a severe collapse in speaker diversity, trailing far behind the other models.

\begin{figure*}[t]
\centering
\begin{tikzpicture}
\pgfplotsset{
  scat-panel/.style={
    width=0.49\linewidth, height=5.2cm,
    xmin=-0.01, xmax=1.00,
    xtick={0,0.25,0.5,0.75,1},
    xticklabels={0,.25,.5,.75,1},
    ymin=3,
    xlabel={Per-class recall (controllability)},
    ylabel={Vendi Score (diversity)},
    grid=both, grid style={draw=gray!15},
    tick label style={font=\footnotesize},
    label style={font=\footnotesize},
    title style={font=\bfseries},
  }
}
\begin{groupplot}[
  group style={group size=2 by 1, horizontal sep=1.3cm},
  scat-panel,
]
\nextgroupplot[title={(a) Gender}, ymax=52, xmin=0.76, xmax=1.01, xtick={0.75,0.8,0.85,0.9,0.95,1.0}, xticklabels={.75,.80,.85,.90,.95,1}, legend cell align=left, legend style={at={(0.5,-0.30)}, anchor=north, legend columns=2, draw=none, font=\scriptsize, text=black, column sep=6pt}]
\addlegendimage{only marks, black, mark=o, mark size=2pt} \addlegendentry{female}
\addlegendimage{only marks, black, mark=*, mark size=2pt} \addlegendentry{male}
\addplot[only marks, color=palRed, mark=o, mark size=2pt, error bars/.cd, y dir=both, y explicit] coordinates {(0.975,38.83) += (0,0.629) -= (0,0.637)};
\addplot[only marks, color=palRed, mark=*, mark size=2pt, error bars/.cd, y dir=both, y explicit] coordinates {(0.950,49.09) += (0,0.682) -= (0,0.686)};
\addplot[only marks, color=palDarkBlue, mark=o, mark size=2pt, error bars/.cd, y dir=both, y explicit] coordinates {(0.965,36.52) += (0,0.419) -= (0,0.412)};
\addplot[only marks, color=palDarkBlue, mark=*, mark size=2pt, error bars/.cd, y dir=both, y explicit] coordinates {(0.984,35.28) += (0,0.714) -= (0,0.721)};
\addplot[only marks, color=palLightBlue, mark=o, mark size=2pt, error bars/.cd, y dir=both, y explicit] coordinates {(0.955,30.59) += (0,0.379) -= (0,0.373)};
\addplot[only marks, color=palLightBlue, mark=*, mark size=2pt, error bars/.cd, y dir=both, y explicit] coordinates {(0.969,35.04) += (0,0.484) -= (0,0.490)};
\addplot[only marks, color=palGreen, mark=o, mark size=2pt, error bars/.cd, y dir=both, y explicit] coordinates {(0.881,17.21) += (0,0.313) -= (0,0.318)};
\addplot[only marks, color=palGreen, mark=*, mark size=2pt, error bars/.cd, y dir=both, y explicit] coordinates {(1.000,20.69) += (0,0.393) -= (0,0.396)};
\addplot[only marks, color=palOrange, mark=o, mark size=2pt, error bars/.cd, y dir=both, y explicit] coordinates {(0.961,16.55) += (0,0.342) -= (0,0.343)};
\addplot[only marks, color=palOrange, mark=*, mark size=2pt, error bars/.cd, y dir=both, y explicit] coordinates {(0.945,28.08) += (0,0.500) -= (0,0.498)};
\addplot[only marks, color=palDarkPurple, mark=o, mark size=2pt, error bars/.cd, y dir=both, y explicit] coordinates {(0.975,16.17) += (0,0.404) -= (0,0.402)};
\addplot[only marks, color=palDarkPurple, mark=*, mark size=2pt, error bars/.cd, y dir=both, y explicit] coordinates {(0.776,24.62) += (0,0.501) -= (0,0.502)};
\addplot[only marks, color=palMidPurple, mark=o, mark size=2pt, error bars/.cd, y dir=both, y explicit] coordinates {(0.984,9.26) += (0,0.244) -= (0,0.237)};
\addplot[only marks, color=palMidPurple, mark=*, mark size=2pt, error bars/.cd, y dir=both, y explicit] coordinates {(0.959,7.98) += (0,0.182) -= (0,0.180)};
\nextgroupplot[title={(b) Age}, ymax=60, legend cell align=left, legend style={at={(0.5,-0.30)}, anchor=north, legend columns=4, draw=none, font=\scriptsize, text=black, column sep=6pt}]
\addlegendimage{only marks, black, mark=o, mark size=2pt} \addlegendentry{child}
\addlegendimage{only marks, black, mark=square*, mark size=2pt} \addlegendentry{young adult}
\addlegendimage{only marks, black, mark=triangle*, mark size=2pt} \addlegendentry{middle-aged}
\addlegendimage{only marks, black, mark=diamond*, mark size=2pt} \addlegendentry{elderly}
\draw[gray!60, dashed] (axis cs:0.25,0) -- (axis cs:0.25,65);
\addplot[only marks, color=palRed, mark=o, mark size=2pt, error bars/.cd, y dir=both, y explicit] coordinates {(0.445,25.06) += (0,0.584) -= (0,0.598)};
\addplot[only marks, color=palRed, mark=square*, mark size=2pt, error bars/.cd, y dir=both, y explicit] coordinates {(0.744,55.50) += (0,0.640) -= (0,0.643)};
\addplot[only marks, color=palRed, mark=triangle*, mark size=2pt, error bars/.cd, y dir=both, y explicit] coordinates {(0.504,57.54) += (0,0.645) -= (0,0.676)};
\addplot[only marks, color=palRed, mark=diamond*, mark size=2pt, error bars/.cd, y dir=both, y explicit] coordinates {(0.435,57.76) += (0,0.620) -= (0,0.615)};
\addplot[only marks, color=palDarkBlue, mark=o, mark size=2pt, error bars/.cd, y dir=both, y explicit] coordinates {(0.024,46.72) += (0,0.532) -= (0,0.526)};
\addplot[only marks, color=palDarkBlue, mark=square*, mark size=2pt, error bars/.cd, y dir=both, y explicit] coordinates {(0.444,46.31) += (0,0.575) -= (0,0.571)};
\addplot[only marks, color=palDarkBlue, mark=triangle*, mark size=2pt, error bars/.cd, y dir=both, y explicit] coordinates {(0.377,48.18) += (0,0.584) -= (0,0.590)};
\addplot[only marks, color=palDarkBlue, mark=diamond*, mark size=2pt, error bars/.cd, y dir=both, y explicit] coordinates {(0.233,49.01) += (0,0.495) -= (0,0.491)};
\addplot[only marks, color=palLightBlue, mark=o, mark size=2pt, error bars/.cd, y dir=both, y explicit] coordinates {(0.013,39.47) += (0,0.496) -= (0,0.493)};
\addplot[only marks, color=palLightBlue, mark=square*, mark size=2pt, error bars/.cd, y dir=both, y explicit] coordinates {(0.345,41.13) += (0,0.496) -= (0,0.491)};
\addplot[only marks, color=palLightBlue, mark=triangle*, mark size=2pt, error bars/.cd, y dir=both, y explicit] coordinates {(0.418,41.20) += (0,0.473) -= (0,0.481)};
\addplot[only marks, color=palLightBlue, mark=diamond*, mark size=2pt, error bars/.cd, y dir=both, y explicit] coordinates {(0.252,40.23) += (0,0.528) -= (0,0.527)};
\addplot[only marks, color=palGreen, mark=o, mark size=2pt, error bars/.cd, y dir=both, y explicit] coordinates {(0.346,23.90) += (0,0.496) -= (0,0.496)};
\addplot[only marks, color=palGreen, mark=square*, mark size=2pt, error bars/.cd, y dir=both, y explicit] coordinates {(0.585,16.96) += (0,0.425) -= (0,0.421)};
\addplot[only marks, color=palGreen, mark=triangle*, mark size=2pt, error bars/.cd, y dir=both, y explicit] coordinates {(0.214,27.36) += (0,0.556) -= (0,0.547)};
\addplot[only marks, color=palGreen, mark=diamond*, mark size=2pt, error bars/.cd, y dir=both, y explicit] coordinates {(0.127,26.86) += (0,0.516) -= (0,0.506)};
\addplot[only marks, color=palOrange, mark=o, mark size=2pt, error bars/.cd, y dir=both, y explicit] coordinates {(0.901,5.51) += (0,0.108) -= (0,0.109)};
\addplot[only marks, color=palOrange, mark=square*, mark size=2pt, error bars/.cd, y dir=both, y explicit] coordinates {(0.622,15.28) += (0,0.469) -= (0,0.454)};
\addplot[only marks, color=palOrange, mark=triangle*, mark size=2pt, error bars/.cd, y dir=both, y explicit] coordinates {(0.283,31.39) += (0,0.615) -= (0,0.628)};
\addplot[only marks, color=palOrange, mark=diamond*, mark size=2pt, error bars/.cd, y dir=both, y explicit] coordinates {(0.144,33.63) += (0,0.699) -= (0,0.690)};
\addplot[only marks, color=palDarkPurple, mark=o, mark size=2pt, error bars/.cd, y dir=both, y explicit] coordinates {(0.058,19.70) += (0,0.537) -= (0,0.526)};
\addplot[only marks, color=palDarkPurple, mark=square*, mark size=2pt, error bars/.cd, y dir=both, y explicit] coordinates {(0.762,22.81) += (0,0.487) -= (0,0.490)};
\addplot[only marks, color=palDarkPurple, mark=triangle*, mark size=2pt, error bars/.cd, y dir=both, y explicit] coordinates {(0.367,25.48) += (0,0.523) -= (0,0.522)};
\addplot[only marks, color=palDarkPurple, mark=diamond*, mark size=2pt, error bars/.cd, y dir=both, y explicit] coordinates {(0.210,28.38) += (0,0.537) -= (0,0.551)};
\addplot[only marks, color=palMidPurple, mark=o, mark size=2pt, error bars/.cd, y dir=both, y explicit] coordinates {(0.550,23.71) += (0,0.564) -= (0,0.560)};
\addplot[only marks, color=palMidPurple, mark=square*, mark size=2pt, error bars/.cd, y dir=both, y explicit] coordinates {(0.306,15.57) += (0,0.457) -= (0,0.439)};
\addplot[only marks, color=palMidPurple, mark=triangle*, mark size=2pt, error bars/.cd, y dir=both, y explicit] coordinates {(0.421,12.83) += (0,0.355) -= (0,0.352)};
\addplot[only marks, color=palMidPurple, mark=diamond*, mark size=2pt, error bars/.cd, y dir=both, y explicit] coordinates {(0.275,16.30) += (0,0.422) -= (0,0.417)};
\end{groupplot}
\begin{axis}[hide axis, scale only axis, width=1pt, height=1pt, xmin=0, xmax=1, ymin=0, ymax=1, legend to name=scatLegOne, legend columns=7, legend style={font=\scriptsize, draw=none, column sep=5pt, text=black}]
\addlegendimage{only marks, color=palRed, mark=*, mark size=2pt} \addlegendentry{MOSS-VG}
\addlegendimage{only marks, color=palDarkBlue, mark=*, mark size=2pt} \addlegendentry{Parler-large}
\addlegendimage{only marks, color=palLightBlue, mark=*, mark size=2pt} \addlegendentry{Parler-mini}
\addlegendimage{only marks, color=palGreen, mark=*, mark size=2pt} \addlegendentry{CapSpeech-NAR}
\addlegendimage{only marks, color=palOrange, mark=*, mark size=2pt} \addlegendentry{Qwen3-TTS}
\addlegendimage{only marks, color=palDarkPurple, mark=*, mark size=2pt} \addlegendentry{VoxInstruct}
\addlegendimage{only marks, color=palMidPurple, mark=*, mark size=2pt} \addlegendentry{VoxCPM2}
\addplot[draw=none, forget plot] coordinates {(0,0)};
\end{axis}
\end{tikzpicture}

\centerline{\ref{scatLegOne}}
 
\caption{\textbf{Controllability versus diversity across gender and age conditions.}
Each color denotes an ITTS model, while marker shapes identify specific gender and age values.
The dashed vertical line marks chance-level recall of 0.25 for age classification.
Error bars show 95\% confidence intervals, which are smaller than the markers.}
\label{fig:scat-gender-age}
\end{figure*}
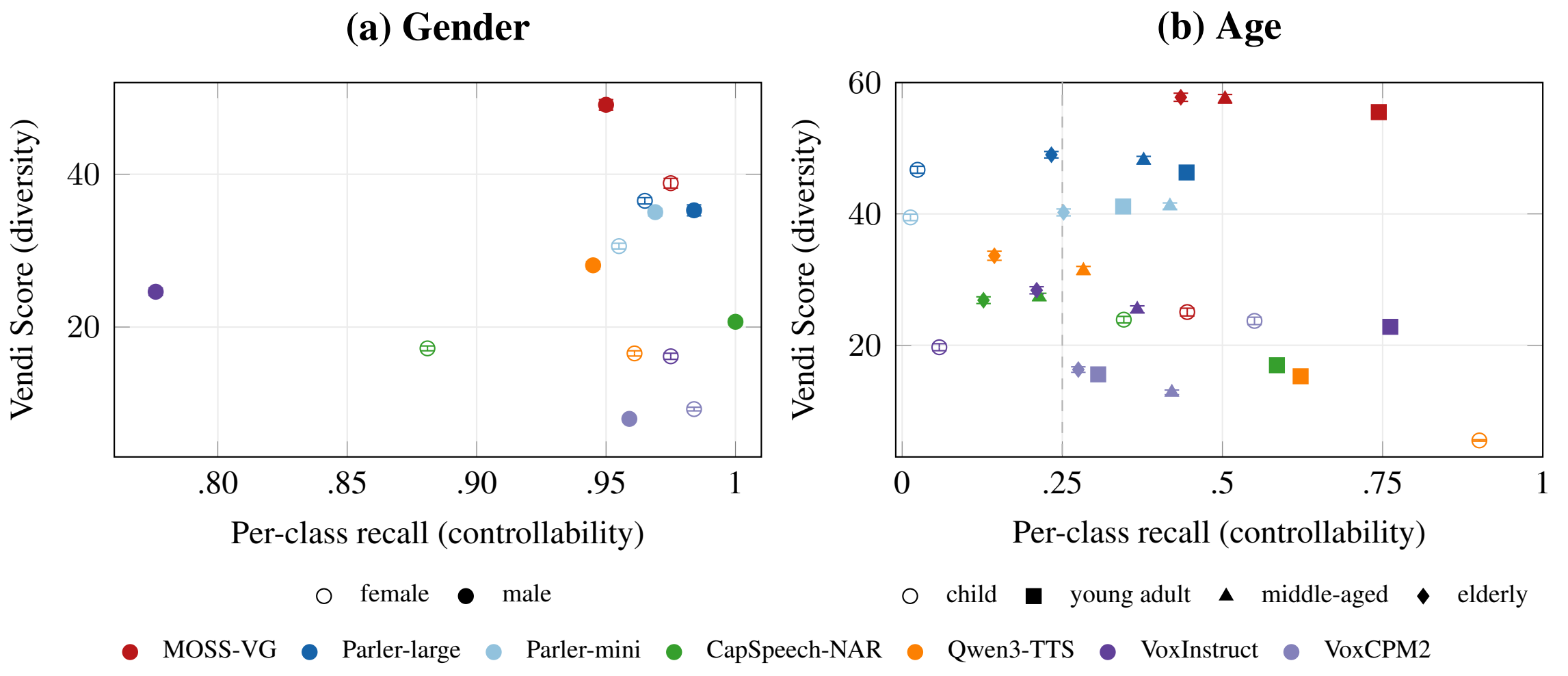

\subsection{Effect of Sampling Parameters}
\label{sec:param-sweep}

To rule out the concern that observed diversity gaps stem from conservative default decoding settings, we sweep sampling parameters across the three systems providing direct decoding controls.
A higher sampling temperature $T$ makes token choices more random and is expected to increase diversity. 
For the classifier-free guidance scale $\gamma$, $\gamma = 1$ corresponds to purely conditional sampling and $\gamma < 1$ moves sampling toward the unconditional prediction, while a larger $\gamma$ pulls outputs closer to the condition and typically trades diversity for fidelity.
Specifically, we vary $T$ for MOSS-VoiceGenerator alongside $\gamma$ for VoxInstruct and VoxCPM2 over $\{0.5, 1.0, 1.5, 2.0, 2.5\}$ across three random seeds.

Figure~\ref{fig:paramsweep} demonstrates that sampling parameters influence speaker diversity. 
For MOSS-VoiceGenerator and VoxCPM2, these changes follow expected trends, with Vendi scores rising as temperature $T$ increases and declining as $\gamma$ exceeds 1, whereas VoxInstruct exhibits no consistent pattern. 
Even under their most diverse configurations, no system matches the real-speech reference across any of the eight encoders. 
Comparing each system at its most diverse setting also preserves the ordering in Figure~\ref{fig:overall}.
For these three systems, the gap to real speech and their relative ranking are therefore not artifacts of the default decoding settings.

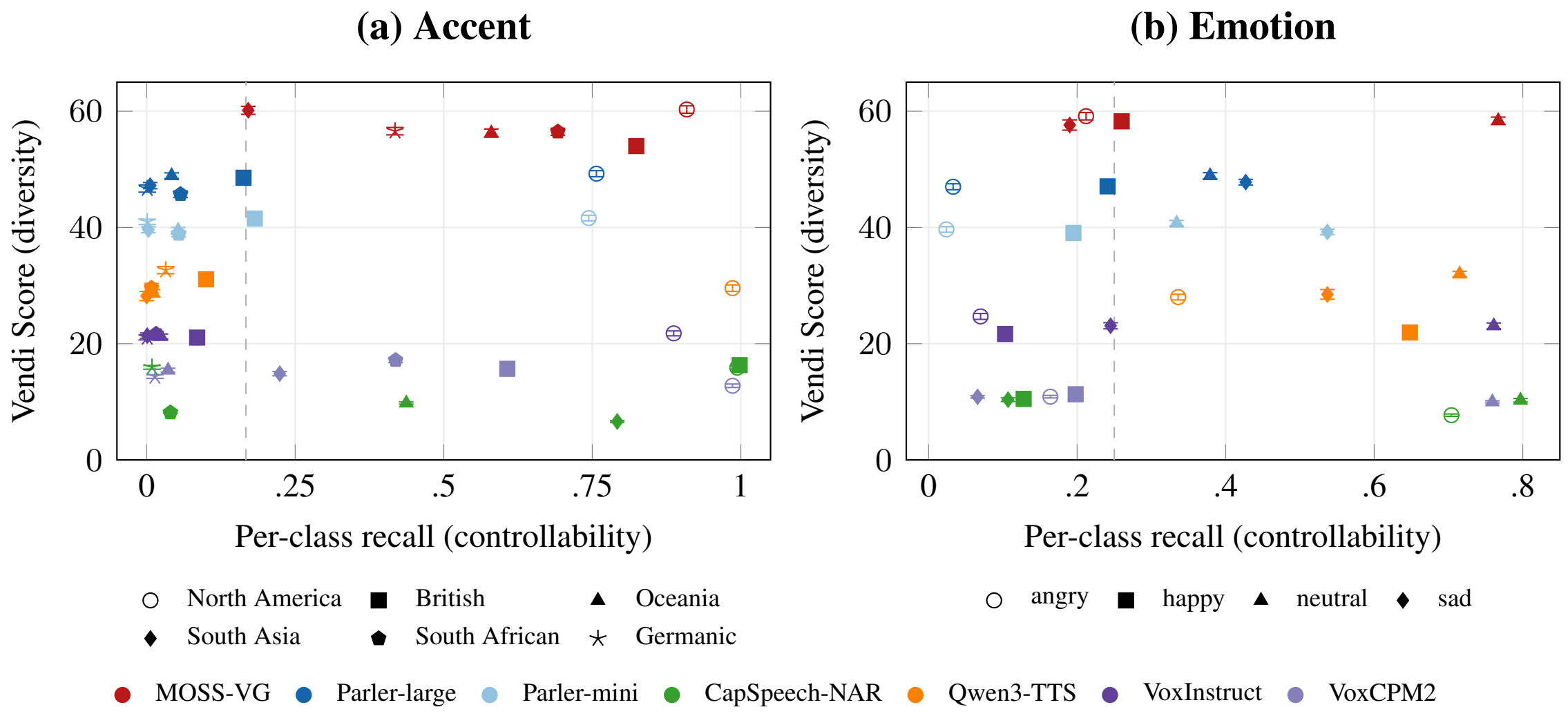
\begin{figure*}[t]
\centering
\begin{tikzpicture}
\pgfplotsset{
  scat-panel/.style={
    width=0.49\linewidth, height=5.2cm,
    xmin=-0.05, xmax=1.05,
    xtick={0,0.25,0.5,0.75,1},
    xticklabels={0,.25,.5,.75,1},
    ymin=0,
    xlabel={Per-class recall (controllability)},
    ylabel={Vendi Score (diversity)},
    grid=both, grid style={draw=gray!15},
    tick label style={font=\footnotesize},
    label style={font=\footnotesize},
    title style={font=\bfseries},
  }
}
\begin{groupplot}[
  group style={group size=2 by 1, horizontal sep=1.3cm},
  scat-panel,
]
\nextgroupplot[title={(a) Accent}, ymax=65, legend cell align=left, legend style={at={(0.5,-0.30)}, anchor=north, legend columns=3, draw=none, font=\scriptsize, text=black, column sep=6pt}]
\addlegendimage{only marks, black, mark=o, mark size=2pt} \addlegendentry{North America}
\addlegendimage{only marks, black, mark=square*, mark size=2pt} \addlegendentry{British}
\addlegendimage{only marks, black, mark=triangle*, mark size=2pt} \addlegendentry{Oceania}
\addlegendimage{only marks, black, mark=diamond*, mark size=2pt} \addlegendentry{South Asia}
\addlegendimage{only marks, black, mark=pentagon*, mark size=2pt} \addlegendentry{South African}
\addlegendimage{only marks, black, mark=star, mark size=2.4pt} \addlegendentry{Germanic}
\draw[gray!60, dashed] (axis cs:0.167,0) -- (axis cs:0.167,65);
\addplot[only marks, color=palRed, mark=o, mark size=2pt, error bars/.cd, y dir=both, y explicit] coordinates {(0.909,60.30) += (0,0.629) -= (0,0.643)};
\addplot[only marks, color=palRed, mark=square*, mark size=2pt, error bars/.cd, y dir=both, y explicit] coordinates {(0.824,54.00) += (0,0.636) -= (0,0.629)};
\addplot[only marks, color=palRed, mark=triangle*, mark size=2pt, error bars/.cd, y dir=both, y explicit] coordinates {(0.580,56.24) += (0,0.694) -= (0,0.713)};
\addplot[only marks, color=palRed, mark=diamond*, mark size=2pt, error bars/.cd, y dir=both, y explicit] coordinates {(0.171,60.16) += (0,0.680) -= (0,0.691)};
\addplot[only marks, color=palRed, mark=pentagon*, mark size=2pt, error bars/.cd, y dir=both, y explicit] coordinates {(0.692,56.47) += (0,0.600) -= (0,0.616)};
\addplot[only marks, color=palRed, mark=star, mark size=2.4pt, error bars/.cd, y dir=both, y explicit] coordinates {(0.418,56.57) += (0,0.633) -= (0,0.643)};
\addplot[only marks, color=palDarkBlue, mark=o, mark size=2pt, error bars/.cd, y dir=both, y explicit] coordinates {(0.757,49.25) += (0,0.523) -= (0,0.520)};
\addplot[only marks, color=palDarkBlue, mark=square*, mark size=2pt, error bars/.cd, y dir=both, y explicit] coordinates {(0.163,48.55) += (0,0.522) -= (0,0.524)};
\addplot[only marks, color=palDarkBlue, mark=triangle*, mark size=2pt, error bars/.cd, y dir=both, y explicit] coordinates {(0.042,48.89) += (0,0.501) -= (0,0.491)};
\addplot[only marks, color=palDarkBlue, mark=diamond*, mark size=2pt, error bars/.cd, y dir=both, y explicit] coordinates {(0.006,47.20) += (0,0.530) -= (0,0.527)};
\addplot[only marks, color=palDarkBlue, mark=pentagon*, mark size=2pt, error bars/.cd, y dir=both, y explicit] coordinates {(0.057,45.72) += (0,0.556) -= (0,0.564)};
\addplot[only marks, color=palDarkBlue, mark=star, mark size=2.4pt, error bars/.cd, y dir=both, y explicit] coordinates {(0.001,46.69) += (0,0.611) -= (0,0.617)};
\addplot[only marks, color=palLightBlue, mark=o, mark size=2pt, error bars/.cd, y dir=both, y explicit] coordinates {(0.744,41.61) += (0,0.477) -= (0,0.474)};
\addplot[only marks, color=palLightBlue, mark=square*, mark size=2pt, error bars/.cd, y dir=both, y explicit] coordinates {(0.182,41.53) += (0,0.475) -= (0,0.476)};
\addplot[only marks, color=palLightBlue, mark=triangle*, mark size=2pt, error bars/.cd, y dir=both, y explicit] coordinates {(0.053,39.50) += (0,0.498) -= (0,0.511)};
\addplot[only marks, color=palLightBlue, mark=diamond*, mark size=2pt, error bars/.cd, y dir=both, y explicit] coordinates {(0.003,39.60) += (0,0.507) -= (0,0.506)};
\addplot[only marks, color=palLightBlue, mark=pentagon*, mark size=2pt, error bars/.cd, y dir=both, y explicit] coordinates {(0.054,38.88) += (0,0.505) -= (0,0.498)};
\addplot[only marks, color=palLightBlue, mark=star, mark size=2.4pt, error bars/.cd, y dir=both, y explicit] coordinates {(0.001,40.96) += (0,0.462) -= (0,0.460)};
\addplot[only marks, color=palGreen, mark=o, mark size=2pt, error bars/.cd, y dir=both, y explicit] coordinates {(0.994,15.88) += (0,0.248) -= (0,0.246)};
\addplot[only marks, color=palGreen, mark=square*, mark size=2pt, error bars/.cd, y dir=both, y explicit] coordinates {(0.998,16.31) += (0,0.281) -= (0,0.276)};
\addplot[only marks, color=palGreen, mark=triangle*, mark size=2pt, error bars/.cd, y dir=both, y explicit] coordinates {(0.437,9.78) += (0,0.200) -= (0,0.198)};
\addplot[only marks, color=palGreen, mark=diamond*, mark size=2pt, error bars/.cd, y dir=both, y explicit] coordinates {(0.792,6.61) += (0,0.158) -= (0,0.157)};
\addplot[only marks, color=palGreen, mark=pentagon*, mark size=2pt, error bars/.cd, y dir=both, y explicit] coordinates {(0.040,8.15) += (0,0.157) -= (0,0.155)};
\addplot[only marks, color=palGreen, mark=star, mark size=2.4pt, error bars/.cd, y dir=both, y explicit] coordinates {(0.009,15.93) += (0,0.325) -= (0,0.319)};
\addplot[only marks, color=palOrange, mark=o, mark size=2pt, error bars/.cd, y dir=both, y explicit] coordinates {(0.986,29.56) += (0,0.550) -= (0,0.567)};
\addplot[only marks, color=palOrange, mark=square*, mark size=2pt, error bars/.cd, y dir=both, y explicit] coordinates {(0.100,31.08) += (0,0.597) -= (0,0.608)};
\addplot[only marks, color=palOrange, mark=triangle*, mark size=2pt, error bars/.cd, y dir=both, y explicit] coordinates {(0.011,28.67) += (0,0.639) -= (0,0.639)};
\addplot[only marks, color=palOrange, mark=diamond*, mark size=2pt, error bars/.cd, y dir=both, y explicit] coordinates {(0.000,28.20) += (0,0.792) -= (0,0.792)};
\addplot[only marks, color=palOrange, mark=pentagon*, mark size=2pt, error bars/.cd, y dir=both, y explicit] coordinates {(0.008,29.59) += (0,0.770) -= (0,0.782)};
\addplot[only marks, color=palOrange, mark=star, mark size=2.4pt, error bars/.cd, y dir=both, y explicit] coordinates {(0.032,32.66) += (0,0.612) -= (0,0.625)};
\addplot[only marks, color=palDarkPurple, mark=o, mark size=2pt, error bars/.cd, y dir=both, y explicit] coordinates {(0.887,21.78) += (0,0.420) -= (0,0.426)};
\addplot[only marks, color=palDarkPurple, mark=square*, mark size=2pt, error bars/.cd, y dir=both, y explicit] coordinates {(0.085,21.05) += (0,0.431) -= (0,0.428)};
\addplot[only marks, color=palDarkPurple, mark=triangle*, mark size=2pt, error bars/.cd, y dir=both, y explicit] coordinates {(0.024,21.23) += (0,0.430) -= (0,0.434)};
\addplot[only marks, color=palDarkPurple, mark=diamond*, mark size=2pt, error bars/.cd, y dir=both, y explicit] coordinates {(0.001,21.38) += (0,0.472) -= (0,0.474)};
\addplot[only marks, color=palDarkPurple, mark=pentagon*, mark size=2pt, error bars/.cd, y dir=both, y explicit] coordinates {(0.016,21.67) += (0,0.439) -= (0,0.440)};
\addplot[only marks, color=palDarkPurple, mark=star, mark size=2.4pt, error bars/.cd, y dir=both, y explicit] coordinates {(0.001,21.08) += (0,0.438) -= (0,0.435)};
\addplot[only marks, color=palMidPurple, mark=o, mark size=2pt, error bars/.cd, y dir=both, y explicit] coordinates {(0.986,12.76) += (0,0.302) -= (0,0.296)};
\addplot[only marks, color=palMidPurple, mark=square*, mark size=2pt, error bars/.cd, y dir=both, y explicit] coordinates {(0.607,15.68) += (0,0.364) -= (0,0.356)};
\addplot[only marks, color=palMidPurple, mark=triangle*, mark size=2pt, error bars/.cd, y dir=both, y explicit] coordinates {(0.036,15.43) += (0,0.349) -= (0,0.354)};
\addplot[only marks, color=palMidPurple, mark=diamond*, mark size=2pt, error bars/.cd, y dir=both, y explicit] coordinates {(0.224,14.84) += (0,0.368) -= (0,0.364)};
\addplot[only marks, color=palMidPurple, mark=pentagon*, mark size=2pt, error bars/.cd, y dir=both, y explicit] coordinates {(0.419,17.16) += (0,0.404) -= (0,0.405)};
\addplot[only marks, color=palMidPurple, mark=star, mark size=2.4pt, error bars/.cd, y dir=both, y explicit] coordinates {(0.014,14.35) += (0,0.323) -= (0,0.314)};
\nextgroupplot[title={(b) Emotion}, ymax=65, xmin=-0.03, xmax=0.85, xtick={0,0.2,0.4,0.6,0.8}, xticklabels={0,.2,.4,.6,.8}, legend cell align=left, legend style={at={(0.5,-0.30)}, anchor=north, legend columns=4, draw=none, font=\scriptsize, text=black, column sep=6pt}]
\addlegendimage{only marks, black, mark=o, mark size=2pt} \addlegendentry{angry}
\addlegendimage{only marks, black, mark=square*, mark size=2pt} \addlegendentry{happy}
\addlegendimage{only marks, black, mark=triangle*, mark size=2pt} \addlegendentry{neutral}
\addlegendimage{only marks, black, mark=diamond*, mark size=2pt} \addlegendentry{sad}
\draw[gray!60, dashed] (axis cs:0.25,0) -- (axis cs:0.25,65);
\addplot[only marks, color=palRed, mark=o, mark size=2pt, error bars/.cd, y dir=both, y explicit] coordinates {(0.212,59.14) += (0,0.622) -= (0,0.639)};
\addplot[only marks, color=palRed, mark=square*, mark size=2pt, error bars/.cd, y dir=both, y explicit] coordinates {(0.260,58.26) += (0,0.750) -= (0,0.760)};
\addplot[only marks, color=palRed, mark=triangle*, mark size=2pt, error bars/.cd, y dir=both, y explicit] coordinates {(0.767,58.36) += (0,0.633) -= (0,0.628)};
\addplot[only marks, color=palRed, mark=diamond*, mark size=2pt, error bars/.cd, y dir=both, y explicit] coordinates {(0.190,57.63) += (0,0.875) -= (0,0.884)};
\addplot[only marks, color=palDarkBlue, mark=o, mark size=2pt, error bars/.cd, y dir=both, y explicit] coordinates {(0.033,47.00) += (0,0.515) -= (0,0.506)};
\addplot[only marks, color=palDarkBlue, mark=square*, mark size=2pt, error bars/.cd, y dir=both, y explicit] coordinates {(0.241,47.07) += (0,0.503) -= (0,0.506)};
\addplot[only marks, color=palDarkBlue, mark=triangle*, mark size=2pt, error bars/.cd, y dir=both, y explicit] coordinates {(0.379,48.89) += (0,0.538) -= (0,0.545)};
\addplot[only marks, color=palDarkBlue, mark=diamond*, mark size=2pt, error bars/.cd, y dir=both, y explicit] coordinates {(0.427,47.80) += (0,0.479) -= (0,0.485)};
\addplot[only marks, color=palLightBlue, mark=o, mark size=2pt, error bars/.cd, y dir=both, y explicit] coordinates {(0.024,39.66) += (0,0.483) -= (0,0.489)};
\addplot[only marks, color=palLightBlue, mark=square*, mark size=2pt, error bars/.cd, y dir=both, y explicit] coordinates {(0.195,39.05) += (0,0.479) -= (0,0.494)};
\addplot[only marks, color=palLightBlue, mark=triangle*, mark size=2pt, error bars/.cd, y dir=both, y explicit] coordinates {(0.334,40.73) += (0,0.469) -= (0,0.473)};
\addplot[only marks, color=palLightBlue, mark=diamond*, mark size=2pt, error bars/.cd, y dir=both, y explicit] coordinates {(0.537,39.23) += (0,0.472) -= (0,0.467)};
\addplot[only marks, color=palGreen, mark=o, mark size=2pt, error bars/.cd, y dir=both, y explicit] coordinates {(0.704,7.70) += (0,0.201) -= (0,0.194)};
\addplot[only marks, color=palGreen, mark=square*, mark size=2pt, error bars/.cd, y dir=both, y explicit] coordinates {(0.128,10.50) += (0,0.281) -= (0,0.276)};
\addplot[only marks, color=palGreen, mark=triangle*, mark size=2pt, error bars/.cd, y dir=both, y explicit] coordinates {(0.797,10.27) += (0,0.307) -= (0,0.308)};
\addplot[only marks, color=palGreen, mark=diamond*, mark size=2pt, error bars/.cd, y dir=both, y explicit] coordinates {(0.107,10.37) += (0,0.311) -= (0,0.312)};
\addplot[only marks, color=palOrange, mark=o, mark size=2pt, error bars/.cd, y dir=both, y explicit] coordinates {(0.336,28.02) += (0,0.481) -= (0,0.493)};
\addplot[only marks, color=palOrange, mark=square*, mark size=2pt, error bars/.cd, y dir=both, y explicit] coordinates {(0.648,21.93) += (0,0.522) -= (0,0.516)};
\addplot[only marks, color=palOrange, mark=triangle*, mark size=2pt, error bars/.cd, y dir=both, y explicit] coordinates {(0.715,31.98) += (0,0.478) -= (0,0.481)};
\addplot[only marks, color=palOrange, mark=diamond*, mark size=2pt, error bars/.cd, y dir=both, y explicit] coordinates {(0.537,28.47) += (0,0.848) -= (0,0.820)};
\addplot[only marks, color=palDarkPurple, mark=o, mark size=2pt, error bars/.cd, y dir=both, y explicit] coordinates {(0.070,24.69) += (0,0.510) -= (0,0.510)};
\addplot[only marks, color=palDarkPurple, mark=square*, mark size=2pt, error bars/.cd, y dir=both, y explicit] coordinates {(0.103,21.67) += (0,0.491) -= (0,0.486)};
\addplot[only marks, color=palDarkPurple, mark=triangle*, mark size=2pt, error bars/.cd, y dir=both, y explicit] coordinates {(0.761,23.06) += (0,0.502) -= (0,0.502)};
\addplot[only marks, color=palDarkPurple, mark=diamond*, mark size=2pt, error bars/.cd, y dir=both, y explicit] coordinates {(0.245,23.09) += (0,0.518) -= (0,0.530)};
\addplot[only marks, color=palMidPurple, mark=o, mark size=2pt, error bars/.cd, y dir=both, y explicit] coordinates {(0.164,10.88) += (0,0.223) -= (0,0.212)};
\addplot[only marks, color=palMidPurple, mark=square*, mark size=2pt, error bars/.cd, y dir=both, y explicit] coordinates {(0.198,11.30) += (0,0.245) -= (0,0.242)};
\addplot[only marks, color=palMidPurple, mark=triangle*, mark size=2pt, error bars/.cd, y dir=both, y explicit] coordinates {(0.759,9.98) += (0,0.189) -= (0,0.181)};
\addplot[only marks, color=palMidPurple, mark=diamond*, mark size=2pt, error bars/.cd, y dir=both, y explicit] coordinates {(0.066,10.85) += (0,0.218) -= (0,0.210)};
\end{groupplot}
\begin{axis}[hide axis, scale only axis, width=1pt, height=1pt, xmin=0, xmax=1, ymin=0, ymax=1, legend to name=scatLegTwo, legend columns=7, legend style={font=\scriptsize, draw=none, column sep=5pt, text=black}]
\addlegendimage{only marks, color=palRed, mark=*, mark size=2pt} \addlegendentry{MOSS-VG}
\addlegendimage{only marks, color=palDarkBlue, mark=*, mark size=2pt} \addlegendentry{Parler-large}
\addlegendimage{only marks, color=palLightBlue, mark=*, mark size=2pt} \addlegendentry{Parler-mini}
\addlegendimage{only marks, color=palGreen, mark=*, mark size=2pt} \addlegendentry{CapSpeech-NAR}
\addlegendimage{only marks, color=palOrange, mark=*, mark size=2pt} \addlegendentry{Qwen3-TTS}
\addlegendimage{only marks, color=palDarkPurple, mark=*, mark size=2pt} \addlegendentry{VoxInstruct}
\addlegendimage{only marks, color=palMidPurple, mark=*, mark size=2pt} \addlegendentry{VoxCPM2}
\addplot[draw=none, forget plot] coordinates {(0,0)};
\end{axis}
\end{tikzpicture}

\centerline{\ref{scatLegTwo}}
 
\caption{\textbf{Controllability versus diversity across accent and emotion conditions.}
Dashed vertical lines mark chance-level recall baselines for each attribute.
Error bars show 95\% confidence intervals, which are smaller than the markers.
}
\label{fig:scat-accent-emotion}
\end{figure*}

\section{Attribute-Conditional Diversity}
\label{sec:conditional}

\subsection{Setup}

For each combination of system and attribute value, we generate 3{,}000 utterances.
Because the Vendi Score scales with the number of utterances, we compute it on $10,000$ random subsets of $N = 1{,}500$ utterances, exactly half of each pool, and report the mean with a 95\% confidence interval.
Since each subset contains half of the pool, scores vary across subsets by approximately as much as the reported mean would vary if the entire pool were regenerated \citep{shao1989jackknife}.
We evaluate the same subsets across all eight encoders, averaging their scores within each subset before computing the interval.
We interpret diversity differences across attribute values through the principle of demographic parity.
Under this principle, repeatedly requesting a female voice should produce just as diverse a pool of voices as requesting a male voice, and we treat any systematic gap between them as a form of bias.
 
\subsection{Gender}
\label{sec:conditional-gender}
 
Figure~\ref{fig:scat-gender-age} (a) illustrates a clear divergence between controllability and diversity. 
First, gender controllability is largely successful across the board. 
Six of the seven evaluated systems achieve per-class recall scores above $0.88$ for both genders, with VoxInstruct being the only exception due to a lower male recall of $0.776$. 
However, strong controllability does not ensure comparable diversity. 
Across five of the seven systems, female speech yields substantially lower Vendi scores than male speech, reaching an extreme in Qwen3-TTS where female diversity drops to $59\%$ of male diversity, scoring $16.6$ compared to $28.1$. 
While models can accurately follow female prompts, they often produce much less diverse voices for female prompts than for male prompts. 
Measuring controllability alone therefore provides an incomplete picture of model behavior.
 
\subsection{Age}
\label{sec:conditional-age}

In Figure~\ref{fig:scat-gender-age}(b), speaker diversity depends on the requested age in five of the seven systems, but barely changes in the two Parler models.
The age group with the lowest diversity differs across these five systems, though child prompts are the most frequent minimum in MOSS-VoiceGenerator, Qwen3-TTS, and VoxInstruct.
For controllability, every system records its lowest recall on either child or elderly prompts, with elderly prompts falling below chance in four systems.
Consequently, age conditioning presents two distinct representational challenges.
Elderly prompts often exhibit low controllability while maintaining high diversity.
In contrast, child prompts can achieve higher controllability while yielding the lowest diversity, as in Qwen3-TTS, where child prompts reach the highest recall of any age group but the lowest Vendi score.

\subsection{Accent}
\label{sec:conditional-accent}

In Figure~\ref{fig:scat-accent-emotion} (a), speaker diversity remains relatively uniform across requested accents, but controllability exhibits a bias favoring North American English.
North American is the only accent that every evaluated ITTS system generates reliably, with per-class recall reaching at least 0.744 across all models.
British English follows as the next strongest accent in most systems, though its recall still lags behind North American in five of the seven models.
By contrast, the remaining four regional varieties suffer severe controllability drops.
Oceanian, South African, and South Asian prompts each fall below the chance baseline in at least four of the seven systems, while Germanic speech fails in six.
Because speaker diversity barely changes with the requested accent, an evaluation based on diversity alone would miss this bias.

\subsection{Emotion}
\label{sec:conditional-emotion}

Figure~\ref{fig:scat-accent-emotion} (b) shows a pattern similar to accent (\S\ref{sec:conditional-accent}).
Speaker diversity remains relatively uniform across prompted emotions in most systems, while controllability favors neutral over expressive emotions.
Recall on neutral prompts exceeds chance in every system and ranks highest among the four emotions in five of the seven systems.
By contrast, angry, happy, and sad prompts each fall below chance recall in at least four systems, with only Qwen3-TTS exceeding chance across all three.
The confusion matrices in Appendix~\ref{app:emotion-confusion} show this breakdown, as speech generated for expressive prompts is frequently classified as neutral, making neutral the most frequent label across all outputs in five systems, while the two Parler models produce sad most often.
High recall on neutral prompts therefore reflects a baseline tendency toward flat, unexpressive acoustics, rather than genuine control over neutral prosody.

\begin{table*}[t]
\centering
\small
\setlength{\tabcolsep}{5pt}
\begin{tabular}{llrrrrrrr}
\toprule
Age & Gender & MOSS-VG & Parler-large & Parler-mini & CapSpeech & Qwen3-TTS & VoxInstruct & VoxCPM2 \\
\midrule
\multirow{2}{*}{Child} & Female & \textbf{27.8} & 36.0 & \textbf{32.0} & 14.2 & \textbf{6.5} & 13.3 & 31.5 \\
 & Male & 43.3 & 35.8 & 36.1 & 17.1 & 25.2 & 21.2 & 38.4 \\
\midrule
\multirow{2}{*}{Young adult} & Female & 36.3 & 36.3 & 32.1 & \textbf{12.4} & 9.5 & \textbf{12.6} & 28.0 \\
 & Male & 46.5 & 36.6 & 35.9 & 13.3 & 19.9 & 17.8 & \textbf{19.9} \\
\midrule
\multirow{2}{*}{Middle-aged} & Female & 40.7 & 35.8 & 32.4 & 19.2 & 24.6 & 12.9 & 23.5 \\
 & Male & 47.3 & \textbf{35.0} & 35.1 & 21.8 & 27.3 & 18.5 & 21.1 \\
\midrule
\multirow{2}{*}{Elderly} & Female & 40.4 & 36.2 & 32.3 & 17.5 & 27.7 & 13.9 & 36.5 \\
 & Male & 48.0 & 36.7 & 35.7 & 19.4 & 25.6 & 20.1 & 30.2 \\
\bottomrule
\end{tabular}
\caption{\textbf{Speaker diversity under joint gender and age conditions.} Vendi scores for joint gender and age prompts, averaged across eight encoders. 
Rows are organized by age group to facilitate female and male comparisons within each bracket. 
Bold numbers highlight the lowest diversity score in each column.}
\label{tab:intersect}
\end{table*}

\begin{table}[t]
\centering
\small
\setlength{\tabcolsep}{3pt}
\begin{tabular}{@{}lcccc}
\toprule
 & \multicolumn{2}{c}{Single attribute} & \multicolumn{2}{c}{Intersectional} \\
\cmidrule(lr){2-3}\cmidrule(l){4-5}
System & Gender & Age & Gender & Age \\
\midrule
MOSS-VG      & female & child       & female & child \\
CapSpeech    & female & young adult & female & young adult \\
Qwen3-TTS    & female & child       & female & child \\
\midrule
VoxInstruct  & female & child       & female & \textbf{young adult}\rlap{$^\dagger$} \\
VoxCPM2      & male   & middle-aged & male   & \textbf{young adult} \\
\midrule
Parler-large & male   & young adult\rlap{$^\dagger$} & male & \textbf{middle-aged}\rlap{$^\dagger$} \\
Parler-mini  & female & child\rlap{$^\dagger$}       & female & child\rlap{$^\dagger$} \\
\bottomrule
\end{tabular}
\caption{\textbf{Gender and age group with the lowest Vendi in each system.} The three row groups are systems where the single-attribute and gender$\times$age results agree, systems where they differ, and systems without a clear lowest age group. \textbf{Bold} marks where the gender$\times$age result differs from the single-attribute result. $^\dagger$Overlaps with the second lowest in confidence interval.}
\label{tab:lowest}
\end{table}

\subsection{Intersectional Fairness: Gender$\times$Age}
\label{sec:conditional-intersect}

Because gender or age exhibit the largest diversity disparity among evaluated attributes, Table~\ref{tab:intersect} examines their intersection by reporting Vendi scores for every combination of gender and age.
Each intersectional condition, such as female child, has 6{,}000 utterances.
This breakdown reveals whether the lower diversity observed for female prompts in \S\ref{sec:conditional-gender} holds across all ages or is confined to specific age groups.
We find that the gender disparity consistently persists across age groups.
Among the five systems where female prompts yield lower diversity overall, four maintain lower female diversity across every age group, namely MOSS-VoiceGenerator, Parler-mini, CapSpeech-NAR, and VoxInstruct.
The remaining model, Qwen3-TTS, follows this pattern across all categories except elderly speakers.

We next examine whether combining a system's least diverse gender and age attributes in a single prompt compounds their individual disadvantages.
Table~\ref{tab:lowest} compares the lowest-diversity attributes identified independently against the lowest intersectional combination.
For MOSS-VoiceGenerator, CapSpeech-NAR, and Qwen3-TTS, the lowest intersectional combination directly pairs their two independently worst attributes.
For VoxInstruct and VoxCPM2, the age group shifts to young adults instead, while for both Parler-TTS models, overlapping confidence intervals leave the minimum statistically ambiguous.
However, scoring lowest can arise simply because the separate penalties of gender and age add together.
We therefore identify compounding only when the gender gap is wider in that age group than in any other.
Under this criterion, only MOSS-VoiceGenerator and Qwen3-TTS exhibit compounding, with female prompts falling furthest behind male prompts specifically among children.
In Qwen3-TTS, this interaction drives female child prompts to the lowest Vendi score in Table~\ref{tab:intersect}.
Across our benchmark, intersectional compounding is therefore model-dependent rather than systematic, emerging in only two of the seven evaluated systems.

\section{Conclusion}

To our knowledge, we provide the first systematic evaluation of ITTS speaker diversity against real speech and across voice attribute prompts.
Across all evaluated systems, generated speech exhibits substantially lower speaker diversity than real recordings.
Furthermore, voice attribute prompts uncover distinct failure modes: while female prompts trigger severe diversity collapses despite accurate gender control, accent and emotion prompts preserve diversity but default heavily to North American English and neutral speech.
These results show that evaluating prompt adherence alone is insufficient, as high controllability does not ensure that a system produces diverse voices.
ITTS should therefore be evaluated on both diversity and controllability, so that synthetic voice does not reinforce biases across voice attributes.

\section*{Limitations}
\label{sec:limitations}
\paragraph{Non-English scope.} Generations, prompts, and content are English only.


\paragraph{Binary voice gender.}
Our evaluation of gender is restricted to prompts for female and male voices, with controllability measured using a binary acoustic classifier.
Both our prompts and classifiers focus on perceived vocal characteristics rather than a speaker's personal gender identity.
Although modern ITTS models can be prompted to generate non-binary or androgynous voices, these fall outside our quantitative analysis because no validated acoustic classifiers or annotated speech corpora currently exist to evaluate them.
This omission reflects a practical measurement constraint rather than a value judgment, and extending diversity audits beyond binary gender remains an important direction for future human perceptual studies.

\paragraph{Closed commercial systems.}
Our evaluation panel focuses exclusively on open-weight and open-source models.
While commercial ITTS platforms such as ElevenLabs Voice Design, Hume Octave, and related proprietary services share instruction-driven interfaces, evaluating them under our full multi-seed sweep is constrained by proprietary access barriers, API usage costs, and limited experimental reproducibility due to opaque server-side model updates.
Benchmarking commercial platforms under standardized protocols represents a valuable direction for future work.

\section*{Ethical Considerations}
\label{sec:ethics}
\paragraph{No human subjects.} All evaluations are on generated audio and the public Emilia speech corpus.
No human subjects were recruited; no IRB approval was required.

\paragraph{Personally identifiable content and data governance.}
All models and evaluators in this study rely on established public speech corpora distributed under academic or open-source licenses, each governed by upstream consent and participant anonymization protocols.
Because synthetic speech is generated purely from descriptive text instructions rather than acoustic reference prompts, the generation process avoids intentional voice cloning or mimicry of specific individuals.
After acceptance, we will release our evaluation code under the MIT License, and our prompt manifests, generated audio, and accent classifier for non-commercial research use.

\paragraph{Environmental impact.} The overall-diversity experiment (\S\ref{sec:overall}) generates 30{,}000 utterances per system, and the attribute-conditional experiment (\S\ref{sec:conditional}) generates 96{,}000 utterances per system, with 3{,}000 for each single-attribute value and 6{,}000 for each gender and age combination. Based on generation speeds measured on NVIDIA B200 GPUs, one complete run of these two experiments takes approximately 842 GPU-hours for generation and 59 GPU-hours for embedding extraction with the eight encoders, or about 901 GPU-hours in total.

\bibliography{custom}

\appendix



\section{Encoder Validation: Cross-Set Metrics and Full Calibration Table}
\label{app:mad}

The diversity metrics in \S\ref{sec:encoder-val}, MPCD and Vendi, evaluate the dispersion of a single set of embeddings in isolation.
Neither metric relies on an external reference population.
A complementary paradigm measures diversity by comparing an evaluated set against a reference population to determine how much of that distribution is covered.
In generative modeling, the recall metric of \citet{kynkaanniemi2019improved} and the coverage metric of \citet{naeem2020reliable} formalize this intuition using nearest-neighbor neighborhoods.
When generated outputs collapse onto a few modes, they leave large portions of the reference distribution uncovered, reflecting a loss of diversity.

We apply this cross-set principle to encoder validation.
Let $R$ denote a reference pool of size $M$ containing utterances from all speakers in the LibriSpeech set from \S\ref{sec:encoder-val}, and let $S$ denote a calibration subset of $N = 240$ utterances drawn from $k$ unique speakers.
Subsets with few speakers cover only a small portion of the reference space, whereas subsets with many speakers span a broader region.
The distance between $S$ and $R$ should therefore decrease as $k$ increases, provided that the encoder maps distinct speakers to separated regions of the embedding space.
We test two cross-set metrics, Chamfer distance and Maximum Mean Discrepancy (MMD), to examine whether population coverage can further inform encoder selection.

The Chamfer distance measures the average nearest-neighbor Euclidean distance between the
two sets, summed over both directions,
\begin{equation}
\begin{split}
\mathrm{CD}(S, R)
&= \frac{1}{|S|} \sum_{x \in S} \min_{y \in R} \lVert x - y \rVert_2 \\
&\quad + \frac{1}{|R|} \sum_{y \in R} \min_{x \in S} \lVert x - y \rVert_2 .
\end{split}
\end{equation}

We further evaluate the Maximum Mean Discrepancy (MMD) with a Gaussian radial basis
function kernel $\kappa(x, y) = \exp\!\big(-\lVert x - y \rVert_2^2 / (2\sigma^2)\big)$, where the
bandwidth $\sigma$ is set by the median heuristic ($\sigma^2$ is the median of all pairwise
squared Euclidean distances over $S \cup R$). We use the biased estimator
\begin{equation}
\begin{split}
\mathrm{MMD}^2(S, R)
&= \frac{1}{|S|^2} \sum_{x, x' \in S} \kappa(x, x') \\
&\quad - \frac{2}{|S||R|} \sum_{x \in S,\, y \in R} \kappa(x, y) \\
&\quad + \frac{1}{|R|^2} \sum_{y, y' \in R} \kappa(y, y').
\end{split}
\end{equation}

\begin{figure*}[t]
\centering
\begin{tikzpicture}
\pgfplotsset{
  metric-panel/.style={
    width=7.6cm, height=5.2cm,
    xmode=log,
    xmin=5.5, xmax=260,
    xtick={6,10,20,40,80,240},
    xticklabels={6,10,20,40,80,240},
    tick label style={font=\scriptsize},
    label style={font=\footnotesize},
    title style={font=\footnotesize},
    xlabel={ground-truth $k$ (distinct speakers)},
    grid=both, grid style={draw=gray!15},
    cycle list={
      {color=palRed,         mark=*,        mark size=1.2pt, thick},
      {color=palOrangeRed,   mark=square*,  mark size=1.2pt, thick},
      {color=palOrange,      mark=triangle*,mark size=1.4pt, thick},
      {color=palLightOrange, mark=diamond*, mark size=1.4pt, thick},
      {color=palGreen,       mark=*,        mark size=1.2pt, thick},
      {color=palLightGreen,  mark=square*,  mark size=1.2pt, thick},
      {color=palSalmon,      mark=triangle*,mark size=1.4pt, thick},
      {color=palDarkPurple,  mark=diamond*, mark size=1.4pt, thick, dashed},
      {color=palDarkBlue,    mark=x,        mark size=2pt,   thick, dashed},
      {color=palBlue,        mark=+,        mark size=2pt,   thick, dashed},
      {color=palLightBlue,   mark=o,        mark size=1.4pt, thick, dashed},
    },
  }
}
\begin{groupplot}[
  group style={group size=2 by 1, horizontal sep=1.4cm},
  metric-panel,
]
\nextgroupplot[title={Chamfer ($\downarrow$ with $k$)}, ylabel={Chamfer}]
\addplot coordinates {(6,0.4873) (8,0.4779) (10,0.4714) (12,0.4635) (15,0.4571) (20,0.4459) (24,0.4373) (30,0.4275) (40,0.4148) (48,0.4068) (60,0.4041) (80,0.4018) (120,0.3991) (240,0.397)};
\addplot coordinates {(6,0.4956) (8,0.4876) (10,0.4803) (12,0.4741) (15,0.4683) (20,0.4594) (24,0.4515) (30,0.4428) (40,0.4318) (48,0.4251) (60,0.4223) (80,0.4206) (120,0.4182) (240,0.4165)};
\addplot coordinates {(6,0.4882) (8,0.4783) (10,0.4728) (12,0.4671) (15,0.4617) (20,0.4524) (24,0.4448) (30,0.4365) (40,0.4257) (48,0.4194) (60,0.4166) (80,0.415) (120,0.4127) (240,0.411)};
\addplot coordinates {(6,0.4608) (8,0.4503) (10,0.4434) (12,0.4373) (15,0.4309) (20,0.4214) (24,0.4133) (30,0.4045) (40,0.3933) (48,0.3866) (60,0.3837) (80,0.3819) (120,0.3794) (240,0.3775)};
\addplot coordinates {(6,0.4601) (8,0.4496) (10,0.4429) (12,0.4362) (15,0.4296) (20,0.4198) (24,0.4115) (30,0.4024) (40,0.3906) (48,0.3834) (60,0.3808) (80,0.3788) (120,0.3762) (240,0.3742)};
\addplot coordinates {(6,0.4379) (8,0.4285) (10,0.4224) (12,0.4159) (15,0.4101) (20,0.401) (24,0.3934) (30,0.3848) (40,0.3745) (48,0.368) (60,0.3653) (80,0.3635) (120,0.3613) (240,0.3594)};
\addplot coordinates {(6,0.3498) (8,0.3394) (10,0.3337) (12,0.327) (15,0.3216) (20,0.3122) (24,0.3062) (30,0.2993) (40,0.2901) (48,0.2842) (60,0.2825) (80,0.281) (120,0.2783) (240,0.2769)};
\addplot coordinates {(6,0.2801) (8,0.272) (10,0.2697) (12,0.265) (15,0.2612) (20,0.2551) (24,0.2503) (30,0.2462) (40,0.2402) (48,0.237) (60,0.2354) (80,0.2344) (120,0.2325) (240,0.2317)};
\addplot coordinates {(6,0.03545) (8,0.03447) (10,0.03403) (12,0.03338) (15,0.03294) (20,0.03227) (24,0.03186) (30,0.03129) (40,0.03069) (48,0.03041) (60,0.03024) (80,0.0301) (120,0.02989) (240,0.02976)};
\addplot coordinates {(6,0.06238) (8,0.05699) (10,0.05545) (12,0.05337) (15,0.0516) (20,0.04909) (24,0.04786) (30,0.04619) (40,0.04474) (48,0.04406) (60,0.04339) (80,0.0429) (120,0.04223) (240,0.04171)};
\addplot coordinates {(6,0.04899) (8,0.04448) (10,0.04353) (12,0.04161) (15,0.03998) (20,0.0379) (24,0.0371) (30,0.03582) (40,0.03475) (48,0.03414) (60,0.03365) (80,0.03328) (120,0.03271) (240,0.03239)};
\nextgroupplot[title={MMD ($\downarrow$ with $k$)}, ylabel={MMD},
  legend to name=metriclegend,
  legend columns=11,
  legend style={font=\tiny, draw=none, column sep=3pt},
]
\addplot coordinates {(6,0.2761) (8,0.2337) (10,0.2068) (12,0.1898) (15,0.1666) (20,0.1411) (24,0.1256) (30,0.1075) (40,0.0878) (48,0.07499) (60,0.06121) (80,0.04048) (120,0.0005085) (240,0)}; \addlegendentry{TIT}
\addplot coordinates {(6,0.26) (8,0.2206) (10,0.1953) (12,0.1782) (15,0.1575) (20,0.1334) (24,0.1184) (30,0.1014) (40,0.08193) (48,0.07105) (60,0.05729) (80,0.03767) (120,0.0004869) (240,0)}; \addlegendentry{SBRN}
\addplot coordinates {(6,0.2596) (8,0.2208) (10,0.1946) (12,0.1783) (15,0.1571) (20,0.1332) (24,0.1175) (30,0.1011) (40,0.08193) (48,0.07116) (60,0.05726) (80,0.03863) (120,0.0004666) (240,0)}; \addlegendentry{SBEC}
\addplot coordinates {(6,0.268) (8,0.226) (10,0.1996) (12,0.1827) (15,0.1618) (20,0.137) (24,0.1221) (30,0.1041) (40,0.08431) (48,0.07263) (60,0.05892) (80,0.03946) (120,0.001205) (240,0)}; \addlegendentry{ER2}
\addplot coordinates {(6,0.2709) (8,0.229) (10,0.2023) (12,0.1842) (15,0.1634) (20,0.1381) (24,0.1227) (30,0.1052) (40,0.08538) (48,0.07371) (60,0.05961) (80,0.03992) (120,0.0006942) (240,0)}; \addlegendentry{WS221}
\addplot coordinates {(6,0.2679) (8,0.226) (10,0.2002) (12,0.1839) (15,0.1627) (20,0.1375) (24,0.1221) (30,0.1044) (40,0.08405) (48,0.07294) (60,0.05912) (80,0.03959) (120,0.0006031) (240,0)}; \addlegendentry{CAM}
\addplot coordinates {(6,0.2744) (8,0.2322) (10,0.2062) (12,0.1893) (15,0.1658) (20,0.1403) (24,0.1256) (30,0.1063) (40,0.08723) (48,0.0752) (60,0.06046) (80,0.04123) (120,0.001182) (240,0)}; \addlegendentry{WSVB}
\addplot coordinates {(6,0.2613) (8,0.2213) (10,0.197) (12,0.182) (15,0.1568) (20,0.1344) (24,0.1201) (30,0.1025) (40,0.08314) (48,0.0713) (60,0.0586) (80,0.03932) (120,0.002295) (240,0)}; \addlegendentry{ER2v2}
\addplot coordinates {(6,0.2593) (8,0.2149) (10,0.1925) (12,0.1779) (15,0.151) (20,0.1286) (24,0.12) (30,0.09733) (40,0.07956) (48,0.06763) (60,0.05623) (80,0.03672) (120,0.006234) (240,0)}; \addlegendentry{XV}
\addplot coordinates {(6,0.2755) (8,0.2237) (10,0.2084) (12,0.1911) (15,0.1633) (20,0.1392) (24,0.1285) (30,0.09977) (40,0.08609) (48,0.07145) (60,0.05813) (80,0.03931) (120,0.008553) (240,0)}; \addlegendentry{WLM}
\addplot coordinates {(6,0.2719) (8,0.2239) (10,0.209) (12,0.1925) (15,0.1627) (20,0.1377) (24,0.13) (30,0.1004) (40,0.08498) (48,0.07034) (60,0.05632) (80,0.03892) (120,0.008178) (240,0)}; \addlegendentry{USAT}
\end{groupplot}
\end{tikzpicture}
 
\vspace{2pt}
\centerline{\ref{metriclegend}}
 
\caption{\textbf{Encoder validation with Chamfer distance and MMD.} Mean Chamfer distance and MMD between 100 random subsets and the full reference pool at each ground-truth speaker count $k$, for all 11 candidate encoders. Solid lines mark the eight encoders retained for evaluation, and dashed lines mark the three excluded encoders.
}
\label{fig:metriccurves-appx}
\end{figure*}
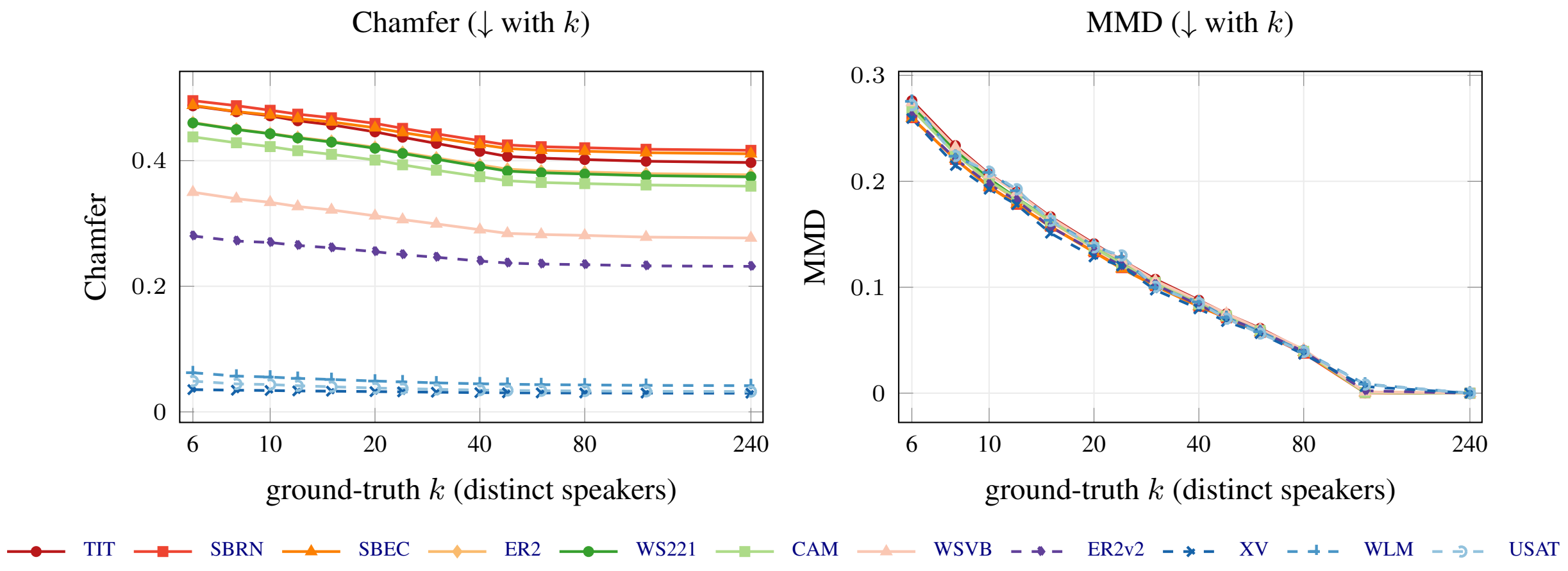

While both cross-set distances reliably track the expansion of subsets toward the reference population, they provide limited discriminatory power for screening candidate encoders.
As illustrated in Figure~\ref{fig:metriccurves-appx}, MMD curves converge to nearly identical trajectories across all architectures, reflecting sample size convergence rather than embedding quality.
Chamfer distance recovers the expected monotonicity but largely mirrors the within-set ranking of MPCD without revealing new failure modes.
Consequently, we use within-set metrics for encoder selection and report these cross-set distances only in this appendix.

Table~\ref{tab:encoders} summarizes the Spearman rank correlation $\rho$ between metric estimates and ground-truth $k$ across all four measures. For each encoder, we compute $\rho$ across all 1{,}400 individual subsets (14 values of $k$ with 100 subsets each). Unlike correlating the 14 mean values, this subset-level evaluation directly penalizes rank inversions when sampling variance causes adjacent speaker counts.

\begin{table}[h]
\centering
\small
\setlength{\tabcolsep}{5pt}
\renewcommand{\arraystretch}{1.15}
\begin{tabular}{lcccc}
\toprule
Encoder & MPCD & VS & chamfer & MMD \\
\midrule
TIT   & \cellcolor{blue!45}0.97 & \cellcolor{blue!55}1.00 & \cellcolor{blue!41}0.96 & \cellcolor{blue!52}0.99 \\
SBRN  & \cellcolor{blue!41}0.96 & \cellcolor{blue!55}1.00 & \cellcolor{blue!31}0.93 & \cellcolor{blue!52}0.99 \\
SBEC  & \cellcolor{blue!34}0.94 & \cellcolor{blue!55}1.00 & \cellcolor{blue!38}0.95 & \cellcolor{blue!52}0.99 \\
ER2   & \cellcolor{red!19}0.79  & \cellcolor{blue!52}0.99 & \cellcolor{blue!38}0.95 & \cellcolor{blue!52}0.99 \\
WS221 & \cellcolor{blue!41}0.96 & \cellcolor{blue!55}1.00 & \cellcolor{blue!38}0.95 & \cellcolor{blue!52}0.99 \\
CAM   & \cellcolor{red!30}0.74  & \cellcolor{blue!52}0.99 & \cellcolor{blue!38}0.95 & \cellcolor{blue!48}0.98 \\
WSVB  & \cellcolor{blue!38}0.95 & \cellcolor{blue!55}1.00 & \cellcolor{blue!41}0.96 & \cellcolor{blue!52}0.99 \\
ER2v2 & \cellcolor{red!21}0.78  & \cellcolor{blue!52}0.99 & \cellcolor{blue!38}0.95 & \cellcolor{blue!52}0.99 \\
\midrule
XV    & \cellcolor{red!59}0.58  & \cellcolor{blue!34}0.94 & \cellcolor{blue!34}0.94 & \cellcolor{blue!41}0.96 \\
WLM   & \cellcolor{red!80}0.44  & \cellcolor{blue!14}0.88 & \cellcolor{blue!41}0.96 & \cellcolor{blue!31}0.93 \\
USAT  & \cellcolor{red!88}0.38  & \cellcolor{red!6}0.83   & \cellcolor{blue!41}0.96 & \cellcolor{blue!24}0.91 \\
\bottomrule
\end{tabular}
\caption{Spearman rank calibration $\rho$ across all four metrics for each evaluated encoder.
Higher values indicate stronger monotonic agreement between metric scores and ground-truth speaker counts $k$.
The top partition contains the eight validated encoders adopted for our main analysis.
The lower partition contains XV, WLM, and USAT.
Chamfer and MMD achieve high correlations of at least 0.91  across all architectures, offering limited sensitivity for encoder selection.}
\label{tab:encoders}
\end{table}

\section{Controllability Classifiers}
\label{app:classifiers}
 
This appendix expands on the four classifiers introduced in \S\ref{sec:controllability}. Three (gender, age, emotion) are pretrained heads used without modification; the fourth (accent) is trained in-house because no publicly released accent classifier covers our six target regions with usable performance. The accent classifier's training setup and validation are reported in Appendix~\ref{app:accent-classifier}.
 
\begin{table*}[h]
\centering
\small
\setlength{\tabcolsep}{4pt}
\begin{tabular}{llc}
\toprule
Attribute & Classifier & Classes \\
\midrule
Gender  & Vox-Profile WavLM sex head & 2 \\
Age     & Vox-Profile WavLM age regressor, binned at $20/40/60$ & 4 \\
Emotion & \texttt{emotion2vec\_plus\_large}, restricted 4-way argmax & 4 \\
Accent  & In-house WavLM-Large + Whisper-Large-V3 LoRA ensemble & 6 \\
\bottomrule
\end{tabular}
\caption{\textbf{Classifiers for measuring controllability.} The classifier and number of classes used for each attribute.}
\label{tab:controllability}
\end{table*}
 
\paragraph{Gender.}
We use the binary sex head from the Vox-Profile \texttt{tiantiaf/wavlm-large-age-sex} model.\footnote{\url{https://huggingface.co/tiantiaf/wavlm-large-age-sex}}
 
\paragraph{Age.}
We use the age regression head from the same \texttt{tiantiaf/wavlm-large-age-sex} model. The head outputs a sigmoid in $[0, 1]$ interpreted as age/100, and we bin the output at 20/40/60 years to recover our four age labels (child, young adult, middle-aged, elderly).
 
\paragraph{Emotion.}
We use the pretrained \texttt{emotion2vec\_plus\_large} model.\footnote{\url{https://github.com/ddlBoJack/emotion2vec}} It predicts nine classes (angry, disgusted, fearful, happy, neutral, other, sad, surprised, and an unknown class). Because our prompts request only four, we take the argmax restricted to the four requested classes.
 
\paragraph{Accent.}
We train a 6-class accent classifier because no released accent head covers our six target regions with usable performance. The classifier is a softmax-averaged ensemble of a WavLM-Large and a Whisper-Large-V3 encoder \cite{radford2023whisper}, both frozen except for LoRA \cite{hu2022lora} rank 16 on the upper-half feed-forward layers. See Appendix~\ref{app:accent-classifier} for training data, recipe, and validation.
 
\section{Accent Classifier Training and Validation}
\label{app:accent-classifier}
 
\paragraph{Training data.}
Table~\ref{tab:accent-train} summarizes the regional speech corpora and corresponding volume used to train the accent classifier.
CommonVoice-17 (CV17) \cite{ardila2020commonvoice} serves as the primary data source across all six target regions.
To address regional data sparsity and enhance acoustic diversity, we supplement three regions with specialized corpora, incorporating L2-ARCTIC \cite{l2arctic} Hindi for South Asia, AfriSpeech-200 \cite{afrispeech} ZA for South Africa, and VoxPopuli \cite{voxpopuli} for Germanic English.
To ensure speaker isolation, we partition each corpus into train, validation, and test splits (60/20/20) by unique speaker identity, ensuring that no speaker appears across multiple subsets.
Furthermore, we restrict each speaker to at most 150 clips so that no individual dominates the training data for any accent.
 
\begin{table*}[h]
\centering
\small
\setlength{\tabcolsep}{3pt}
\begin{tabular}{llrr}
\toprule
Region & Source(s) & Train clips (speakers) & Hours \\
\midrule
North America   & CV17                                            & 40{,}000 (4{,}350) & 58.6 \\
British English & CV17                                            & 20{,}000 (1{,}410) & 29.0 \\
South Asia      & CV17 + L2-ARCTIC Hindi                          & 16{,}000 (1{,}221) & 24.8 \\
Oceania         & CV17                                            & 14{,}000 (548)   & 20.6 \\
South African   & AfriSpeech-200 ZA + CV17                        &  8{,}495 (313)   & 15.4 \\
Germanic        & CV17 + VoxPopuli \texttt{en\_de}/\texttt{en\_nl} &  1{,}799 (109)   &  3.5 \\
\bottomrule
\end{tabular}
\caption{\textbf{Accent classifier training data.} CV17 denotes CommonVoice-17. VoxPopuli \texttt{en\_de}/\texttt{en\_nl} adds 75 Germanic speakers (999 clips) from the European Parliament channel. AfriSpeech-200 \texttt{ZA} adds 119 South-African speakers (2{,}793 clips). L2-ARCTIC Hindi adds 76 clips (2 speakers) to South Asia.}
\label{tab:accent-train}
\end{table*}
 
\paragraph{Architecture and training recipe.}
The classification model operates as an ensemble combining two distinct acoustic foundation backbones, WavLM-Large and Whisper-Large-V3.
Each backbone remains frozen, with parameter-efficient finetuning applied through LoRA (rank 16) inserted into the upper-half feed-forward projections, yielding approximately 2.26M trainable parameters for the WavLM model (0.71\% of its parameters). 
The WavLM model aggregates the hidden representations of all transformer layers via a learnable softmax-weighted average, while the Whisper model uses the output of its last encoder layer, following the Vox-Profile implementation. 
Each is followed by three pointwise 1D convolutional layers, temporal average pooling, and a two-layer linear classification head producing six-way logits.
Hidden representations across all transformer layers are aggregated via a learnable softmax-weighted average, followed by three pointwise 1D convolutional layers, temporal average pooling, and a two-layer linear classification head producing six-way logits.
Final predictions are obtained by averaging the softmax posteriors of both backbones.

We train the model using the Adam optimizer with a constant learning rate of $5\times 10^{-4}$ for 15 epochs, setting the effective batch size to 16 with mixed-precision training.
To counteract the class imbalance shown in Table~\ref{tab:accent-train}, we optimize an inverse-frequency weighted cross-entropy loss.
Data augmentation during training includes additive Gaussian noise at SNR between 3 dB and 30 dB, random time masking covering 10\% to 15\% of total duration, time stretching by factors between 0.9 and 1.1, and polarity inversion.
The final model checkpoint is selected based on maximum validation macro-F1.
 
\paragraph{Validation.}
The validation results presented in Table~\ref{tab:accent-val} confirm that the classifier provides a sound evaluator for the controllability analysis in \S\ref{sec:controllability}.
First, the overall macro-F1 of $0.876$ substantially exceeds the random chance level of $0.167$, demonstrating reliable regional discrimination across all target categories.
Second, per-class F1 scores remain tightly clustered between $0.825$ and $0.918$, ensuring that cross-region controllability differences cannot be attributed to uneven classifier sensitivity.
Even the lower-resource Germanic category attains an F1 of $0.887$, performing on par with well-represented regions such as North America and South Africa.
 
\begin{table}[h]
\centering
\small
\setlength{\tabcolsep}{6pt}
\begin{tabular}{lccc}
\toprule
Region & WavLM & Whisper & Ensemble \\
\midrule
North America   & 0.868          & 0.880          & \textbf{0.881} \\
British English & 0.802          & 0.771          & \textbf{0.825} \\
South Asia      & 0.905          & \textbf{0.920} & 0.918 \\
Germanic        & 0.842          & \textbf{0.906} & 0.887 \\
Oceania         & 0.840          & 0.793          & \textbf{0.873} \\
South African   & 0.832          & 0.869          & \textbf{0.871} \\
\midrule
Macro-F1        & 0.848          & 0.856          & \textbf{0.876} \\
\bottomrule
\end{tabular}
\caption{Per-class F1 on the speaker-independent held-out test split. The ensemble (production model used for controllability measurement in \S\ref{sec:controllability}) achieves the best macro-F1; on per-class F1 Whisper edges the ensemble on South Asia and Germanic by small margins.}
\label{tab:accent-val}
\end{table}

\section{Emotion Confusion Matrices}
\label{app:emotion-confusion}
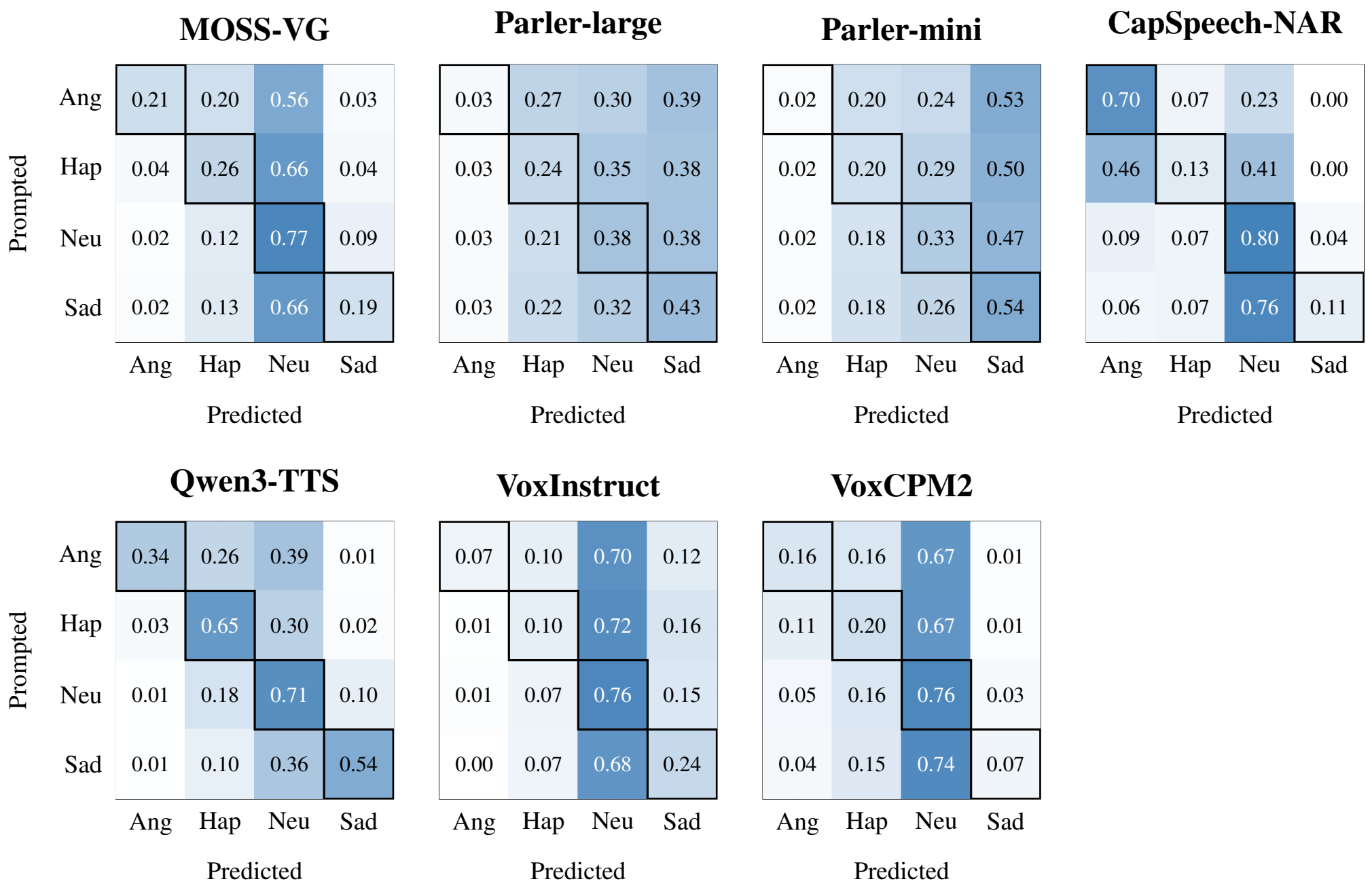
\begin{figure*}[h]
\centering
\begin{tikzpicture}
\begin{groupplot}[
  group style={group size=4 by 2, horizontal sep=0.45cm, vertical sep=1.75cm, yticklabels at=edge left, ylabels at=edge left},
  scale only axis, width=2.7cm, height=2.7cm,
  xmin=-0.5, xmax=3.5, ymin=-0.5, ymax=3.5, y dir=reverse,
  xtick={0,1,2,3}, ytick={0,1,2,3},
  xticklabels={Ang,Hap,Neu,Sad}, yticklabels={Ang,Hap,Neu,Sad},
  tick label style={font=\scriptsize}, major tick length=0pt,
  xlabel={Predicted}, ylabel={Prompted}, label style={font=\scriptsize},
  title style={font=\footnotesize\bfseries, yshift=-3pt},
  colormap={wb}{color(0cm)=(white); color(1cm)=(palDarkBlue)},
  point meta min=0, point meta max=1,
]
\nextgroupplot[title={MOSS-VG}]
\addplot[matrix plot*, mesh/cols=4, point meta=explicit] coordinates {(0,0) [0.2120] (1,0) [0.2050] (2,0) [0.5550] (3,0) [0.0280] (0,1) [0.0407] (1,1) [0.2603] (2,1) [0.6557] (3,1) [0.0433] (0,2) [0.0153] (1,2) [0.1247] (2,2) [0.7673] (3,2) [0.0927] (0,3) [0.0177] (1,3) [0.1343] (2,3) [0.6577] (3,3) [0.1903]};
\node[font=\tiny, text=black] at (axis cs:0,0) {0.21};
\node[font=\tiny, text=black] at (axis cs:1,0) {0.20};
\node[font=\tiny, text=white] at (axis cs:2,0) {0.56};
\node[font=\tiny, text=black] at (axis cs:3,0) {0.03};
\node[font=\tiny, text=black] at (axis cs:0,1) {0.04};
\node[font=\tiny, text=black] at (axis cs:1,1) {0.26};
\node[font=\tiny, text=white] at (axis cs:2,1) {0.66};
\node[font=\tiny, text=black] at (axis cs:3,1) {0.04};
\node[font=\tiny, text=black] at (axis cs:0,2) {0.02};
\node[font=\tiny, text=black] at (axis cs:1,2) {0.12};
\node[font=\tiny, text=white] at (axis cs:2,2) {0.77};
\node[font=\tiny, text=black] at (axis cs:3,2) {0.09};
\node[font=\tiny, text=black] at (axis cs:0,3) {0.02};
\node[font=\tiny, text=black] at (axis cs:1,3) {0.13};
\node[font=\tiny, text=white] at (axis cs:2,3) {0.66};
\node[font=\tiny, text=black] at (axis cs:3,3) {0.19};
\draw[black, line width=0.6pt] (axis cs:-0.5,-0.5) rectangle (axis cs:0.5,0.5);
\draw[black, line width=0.6pt] (axis cs:0.5,0.5) rectangle (axis cs:1.5,1.5);
\draw[black, line width=0.6pt] (axis cs:1.5,1.5) rectangle (axis cs:2.5,2.5);
\draw[black, line width=0.6pt] (axis cs:2.5,2.5) rectangle (axis cs:3.5,3.5);
\nextgroupplot[title={Parler-large}]
\addplot[matrix plot*, mesh/cols=4, point meta=explicit] coordinates {(0,0) [0.0333] (1,0) [0.2740] (2,0) [0.2997] (3,0) [0.3930] (0,1) [0.0317] (1,1) [0.2413] (2,1) [0.3520] (3,1) [0.3750] (0,2) [0.0300] (1,2) [0.2087] (2,2) [0.3790] (3,2) [0.3823] (0,3) [0.0313] (1,3) [0.2240] (2,3) [0.3177] (3,3) [0.4270]};
\node[font=\tiny, text=black] at (axis cs:0,0) {0.03};
\node[font=\tiny, text=black] at (axis cs:1,0) {0.27};
\node[font=\tiny, text=black] at (axis cs:2,0) {0.30};
\node[font=\tiny, text=black] at (axis cs:3,0) {0.39};
\node[font=\tiny, text=black] at (axis cs:0,1) {0.03};
\node[font=\tiny, text=black] at (axis cs:1,1) {0.24};
\node[font=\tiny, text=black] at (axis cs:2,1) {0.35};
\node[font=\tiny, text=black] at (axis cs:3,1) {0.38};
\node[font=\tiny, text=black] at (axis cs:0,2) {0.03};
\node[font=\tiny, text=black] at (axis cs:1,2) {0.21};
\node[font=\tiny, text=black] at (axis cs:2,2) {0.38};
\node[font=\tiny, text=black] at (axis cs:3,2) {0.38};
\node[font=\tiny, text=black] at (axis cs:0,3) {0.03};
\node[font=\tiny, text=black] at (axis cs:1,3) {0.22};
\node[font=\tiny, text=black] at (axis cs:2,3) {0.32};
\node[font=\tiny, text=black] at (axis cs:3,3) {0.43};
\draw[black, line width=0.6pt] (axis cs:-0.5,-0.5) rectangle (axis cs:0.5,0.5);
\draw[black, line width=0.6pt] (axis cs:0.5,0.5) rectangle (axis cs:1.5,1.5);
\draw[black, line width=0.6pt] (axis cs:1.5,1.5) rectangle (axis cs:2.5,2.5);
\draw[black, line width=0.6pt] (axis cs:2.5,2.5) rectangle (axis cs:3.5,3.5);
\nextgroupplot[title={Parler-mini}]
\addplot[matrix plot*, mesh/cols=4, point meta=explicit] coordinates {(0,0) [0.0243] (1,0) [0.2050] (2,0) [0.2360] (3,0) [0.5347] (0,1) [0.0193] (1,1) [0.1950] (2,1) [0.2853] (3,1) [0.5003] (0,2) [0.0213] (1,2) [0.1780] (2,2) [0.3340] (3,2) [0.4667] (0,3) [0.0187] (1,3) [0.1830] (2,3) [0.2610] (3,3) [0.5373]};
\node[font=\tiny, text=black] at (axis cs:0,0) {0.02};
\node[font=\tiny, text=black] at (axis cs:1,0) {0.20};
\node[font=\tiny, text=black] at (axis cs:2,0) {0.24};
\node[font=\tiny, text=black] at (axis cs:3,0) {0.53};
\node[font=\tiny, text=black] at (axis cs:0,1) {0.02};
\node[font=\tiny, text=black] at (axis cs:1,1) {0.20};
\node[font=\tiny, text=black] at (axis cs:2,1) {0.29};
\node[font=\tiny, text=black] at (axis cs:3,1) {0.50};
\node[font=\tiny, text=black] at (axis cs:0,2) {0.02};
\node[font=\tiny, text=black] at (axis cs:1,2) {0.18};
\node[font=\tiny, text=black] at (axis cs:2,2) {0.33};
\node[font=\tiny, text=black] at (axis cs:3,2) {0.47};
\node[font=\tiny, text=black] at (axis cs:0,3) {0.02};
\node[font=\tiny, text=black] at (axis cs:1,3) {0.18};
\node[font=\tiny, text=black] at (axis cs:2,3) {0.26};
\node[font=\tiny, text=black] at (axis cs:3,3) {0.54};
\draw[black, line width=0.6pt] (axis cs:-0.5,-0.5) rectangle (axis cs:0.5,0.5);
\draw[black, line width=0.6pt] (axis cs:0.5,0.5) rectangle (axis cs:1.5,1.5);
\draw[black, line width=0.6pt] (axis cs:1.5,1.5) rectangle (axis cs:2.5,2.5);
\draw[black, line width=0.6pt] (axis cs:2.5,2.5) rectangle (axis cs:3.5,3.5);
\nextgroupplot[title={CapSpeech-NAR}]
\addplot[matrix plot*, mesh/cols=4, point meta=explicit] coordinates {(0,0) [0.7043] (1,0) [0.0697] (2,0) [0.2253] (3,0) [0.0007] (0,1) [0.4630] (1,1) [0.1283] (2,1) [0.4053] (3,1) [0.0033] (0,2) [0.0930] (1,2) [0.0683] (2,2) [0.7973] (3,2) [0.0413] (0,3) [0.0603] (1,3) [0.0677] (2,3) [0.7647] (3,3) [0.1073]};
\node[font=\tiny, text=white] at (axis cs:0,0) {0.70};
\node[font=\tiny, text=black] at (axis cs:1,0) {0.07};
\node[font=\tiny, text=black] at (axis cs:2,0) {0.23};
\node[font=\tiny, text=black] at (axis cs:3,0) {0.00};
\node[font=\tiny, text=black] at (axis cs:0,1) {0.46};
\node[font=\tiny, text=black] at (axis cs:1,1) {0.13};
\node[font=\tiny, text=black] at (axis cs:2,1) {0.41};
\node[font=\tiny, text=black] at (axis cs:3,1) {0.00};
\node[font=\tiny, text=black] at (axis cs:0,2) {0.09};
\node[font=\tiny, text=black] at (axis cs:1,2) {0.07};
\node[font=\tiny, text=white] at (axis cs:2,2) {0.80};
\node[font=\tiny, text=black] at (axis cs:3,2) {0.04};
\node[font=\tiny, text=black] at (axis cs:0,3) {0.06};
\node[font=\tiny, text=black] at (axis cs:1,3) {0.07};
\node[font=\tiny, text=white] at (axis cs:2,3) {0.76};
\node[font=\tiny, text=black] at (axis cs:3,3) {0.11};
\draw[black, line width=0.6pt] (axis cs:-0.5,-0.5) rectangle (axis cs:0.5,0.5);
\draw[black, line width=0.6pt] (axis cs:0.5,0.5) rectangle (axis cs:1.5,1.5);
\draw[black, line width=0.6pt] (axis cs:1.5,1.5) rectangle (axis cs:2.5,2.5);
\draw[black, line width=0.6pt] (axis cs:2.5,2.5) rectangle (axis cs:3.5,3.5);
\nextgroupplot[title={Qwen3-TTS}]
\addplot[matrix plot*, mesh/cols=4, point meta=explicit] coordinates {(0,0) [0.3363] (1,0) [0.2600] (2,0) [0.3893] (3,0) [0.0143] (0,1) [0.0347] (1,1) [0.6477] (2,1) [0.2980] (3,1) [0.0197] (0,2) [0.0087] (1,2) [0.1757] (2,2) [0.7150] (3,2) [0.1007] (0,3) [0.0083] (1,3) [0.0967] (2,3) [0.3577] (3,3) [0.5373]};
\node[font=\tiny, text=black] at (axis cs:0,0) {0.34};
\node[font=\tiny, text=black] at (axis cs:1,0) {0.26};
\node[font=\tiny, text=black] at (axis cs:2,0) {0.39};
\node[font=\tiny, text=black] at (axis cs:3,0) {0.01};
\node[font=\tiny, text=black] at (axis cs:0,1) {0.03};
\node[font=\tiny, text=white] at (axis cs:1,1) {0.65};
\node[font=\tiny, text=black] at (axis cs:2,1) {0.30};
\node[font=\tiny, text=black] at (axis cs:3,1) {0.02};
\node[font=\tiny, text=black] at (axis cs:0,2) {0.01};
\node[font=\tiny, text=black] at (axis cs:1,2) {0.18};
\node[font=\tiny, text=white] at (axis cs:2,2) {0.71};
\node[font=\tiny, text=black] at (axis cs:3,2) {0.10};
\node[font=\tiny, text=black] at (axis cs:0,3) {0.01};
\node[font=\tiny, text=black] at (axis cs:1,3) {0.10};
\node[font=\tiny, text=black] at (axis cs:2,3) {0.36};
\node[font=\tiny, text=black] at (axis cs:3,3) {0.54};
\draw[black, line width=0.6pt] (axis cs:-0.5,-0.5) rectangle (axis cs:0.5,0.5);
\draw[black, line width=0.6pt] (axis cs:0.5,0.5) rectangle (axis cs:1.5,1.5);
\draw[black, line width=0.6pt] (axis cs:1.5,1.5) rectangle (axis cs:2.5,2.5);
\draw[black, line width=0.6pt] (axis cs:2.5,2.5) rectangle (axis cs:3.5,3.5);
\nextgroupplot[title={VoxInstruct}]
\addplot[matrix plot*, mesh/cols=4, point meta=explicit] coordinates {(0,0) [0.0700] (1,0) [0.1033] (2,0) [0.7043] (3,0) [0.1223] (0,1) [0.0087] (1,1) [0.1027] (2,1) [0.7240] (3,1) [0.1647] (0,2) [0.0117] (1,2) [0.0747] (2,2) [0.7607] (3,2) [0.1530] (0,3) [0.0043] (1,3) [0.0717] (2,3) [0.6793] (3,3) [0.2447]};
\node[font=\tiny, text=black] at (axis cs:0,0) {0.07};
\node[font=\tiny, text=black] at (axis cs:1,0) {0.10};
\node[font=\tiny, text=white] at (axis cs:2,0) {0.70};
\node[font=\tiny, text=black] at (axis cs:3,0) {0.12};
\node[font=\tiny, text=black] at (axis cs:0,1) {0.01};
\node[font=\tiny, text=black] at (axis cs:1,1) {0.10};
\node[font=\tiny, text=white] at (axis cs:2,1) {0.72};
\node[font=\tiny, text=black] at (axis cs:3,1) {0.16};
\node[font=\tiny, text=black] at (axis cs:0,2) {0.01};
\node[font=\tiny, text=black] at (axis cs:1,2) {0.07};
\node[font=\tiny, text=white] at (axis cs:2,2) {0.76};
\node[font=\tiny, text=black] at (axis cs:3,2) {0.15};
\node[font=\tiny, text=black] at (axis cs:0,3) {0.00};
\node[font=\tiny, text=black] at (axis cs:1,3) {0.07};
\node[font=\tiny, text=white] at (axis cs:2,3) {0.68};
\node[font=\tiny, text=black] at (axis cs:3,3) {0.24};
\draw[black, line width=0.6pt] (axis cs:-0.5,-0.5) rectangle (axis cs:0.5,0.5);
\draw[black, line width=0.6pt] (axis cs:0.5,0.5) rectangle (axis cs:1.5,1.5);
\draw[black, line width=0.6pt] (axis cs:1.5,1.5) rectangle (axis cs:2.5,2.5);
\draw[black, line width=0.6pt] (axis cs:2.5,2.5) rectangle (axis cs:3.5,3.5);
\nextgroupplot[title={VoxCPM2}]
\addplot[matrix plot*, mesh/cols=4, point meta=explicit] coordinates {(0,0) [0.1637] (1,0) [0.1570] (2,0) [0.6677] (3,0) [0.0117] (0,1) [0.1127] (1,1) [0.1977] (2,1) [0.6747] (3,1) [0.0150] (0,2) [0.0510] (1,2) [0.1560] (2,2) [0.7590] (3,2) [0.0340] (0,3) [0.0437] (1,3) [0.1463] (2,3) [0.7443] (3,3) [0.0657]};
\node[font=\tiny, text=black] at (axis cs:0,0) {0.16};
\node[font=\tiny, text=black] at (axis cs:1,0) {0.16};
\node[font=\tiny, text=white] at (axis cs:2,0) {0.67};
\node[font=\tiny, text=black] at (axis cs:3,0) {0.01};
\node[font=\tiny, text=black] at (axis cs:0,1) {0.11};
\node[font=\tiny, text=black] at (axis cs:1,1) {0.20};
\node[font=\tiny, text=white] at (axis cs:2,1) {0.67};
\node[font=\tiny, text=black] at (axis cs:3,1) {0.01};
\node[font=\tiny, text=black] at (axis cs:0,2) {0.05};
\node[font=\tiny, text=black] at (axis cs:1,2) {0.16};
\node[font=\tiny, text=white] at (axis cs:2,2) {0.76};
\node[font=\tiny, text=black] at (axis cs:3,2) {0.03};
\node[font=\tiny, text=black] at (axis cs:0,3) {0.04};
\node[font=\tiny, text=black] at (axis cs:1,3) {0.15};
\node[font=\tiny, text=white] at (axis cs:2,3) {0.74};
\node[font=\tiny, text=black] at (axis cs:3,3) {0.07};
\draw[black, line width=0.6pt] (axis cs:-0.5,-0.5) rectangle (axis cs:0.5,0.5);
\draw[black, line width=0.6pt] (axis cs:0.5,0.5) rectangle (axis cs:1.5,1.5);
\draw[black, line width=0.6pt] (axis cs:1.5,1.5) rectangle (axis cs:2.5,2.5);
\draw[black, line width=0.6pt] (axis cs:2.5,2.5) rectangle (axis cs:3.5,3.5);
\end{groupplot}
\end{tikzpicture}
\caption{\textbf{Emotion confusion matrices for the seven ITTS systems.} Rows give the prompted emotion and columns the emotion predicted by the classifier, with each row normalized over its 3{,}000 clips. Outlined diagonal cells equal the per-class recall plotted in Figure~\ref{fig:scat-accent-emotion}(b).}
\label{fig:emotion-confusion}
\end{figure*}

Figure~\ref{fig:emotion-confusion} presents the emotion confusion matrices for all seven systems.
Misclassified outputs are most often labeled neutral in five systems and sad in the two Parler models.
In MOSS-VoiceGenerator, VoxInstruct, and VoxCPM2, neutral receives the largest share of predictions in every row, ranging from 0.56 to 0.77 regardless of the prompted emotion.
CapSpeech-NAR renders angry prompts with a recall of 0.70, but its happy prompts are most often classified as angry at 0.46 and its sad prompts as neutral at 0.76.
In Parler-mini, sad receives the largest share in every row, ranging from 0.47 to 0.54, while Parler-large shows a weaker version of this pattern, with sad and neutral nearly tied in its neutral row.
Qwen3-TTS is the only system in which the prompted emotion receives the largest share in three of the four rows, namely happy, neutral, and sad, while angry prompts are classified as neutral at 0.39 slightly more often than as angry at 0.34.
In every system, the emotion that receives the most misclassified outputs is also the emotion with the highest recall.
Recall on that emotion therefore mixes prompt following with the system tendency to produce that emotion under any prompt.

\section{Per-System Speaker Diversity with Confidence Intervals}
\label{app:per-system}
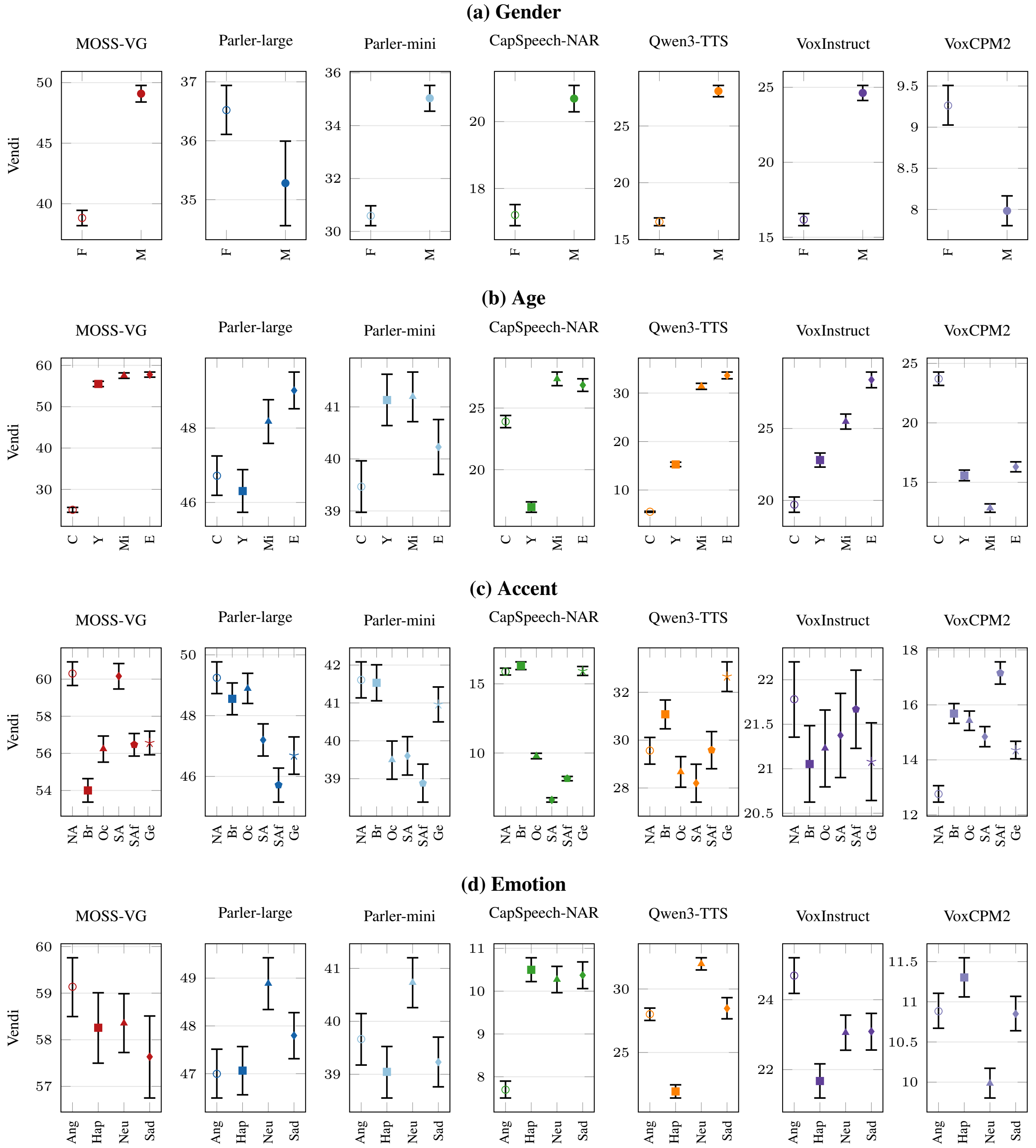
\begin{figure*}[p]
\centering
{\small\textbf{(a) Gender}}\\[1pt]
\begin{tikzpicture}\begin{groupplot}[group style={group size=7 by 1, horizontal sep=0.68cm}, scale only axis, width=1.55cm, height=2.6cm, symbolic x coords={F,M}, xtick={F,M}, enlarge x limits=0.35, tick label style={font=\tiny}, x tick label style={rotate=90, anchor=east}, title style={font=\scriptsize}, ymajorgrids, grid style={draw=gray!20}, yticklabel style={/pgf/number format/fixed, /pgf/number format/precision=1}]
\nextgroupplot[title={MOSS-VG}, ylabel={Vendi}, ylabel style={font=\scriptsize}]
\addplot[only marks, color=palRed, mark=o, mark size=1.6pt, error bars/.cd, y dir=both, y explicit, error bar style={line width=0.7pt, black}, error mark options={black, line width=0.7pt, mark size=2.5pt, rotate=90}] coordinates {(F,38.826) += (0,0.629) -= (0,0.637)};
\addplot[only marks, color=palRed, mark=*, mark size=1.6pt, error bars/.cd, y dir=both, y explicit, error bar style={line width=0.7pt, black}, error mark options={black, line width=0.7pt, mark size=2.5pt, rotate=90}] coordinates {(M,49.086) += (0,0.682) -= (0,0.686)};
\nextgroupplot[title={Parler-large}]
\addplot[only marks, color=palDarkBlue, mark=o, mark size=1.6pt, error bars/.cd, y dir=both, y explicit, error bar style={line width=0.7pt, black}, error mark options={black, line width=0.7pt, mark size=2.5pt, rotate=90}] coordinates {(F,36.520) += (0,0.419) -= (0,0.412)};
\addplot[only marks, color=palDarkBlue, mark=*, mark size=1.6pt, error bars/.cd, y dir=both, y explicit, error bar style={line width=0.7pt, black}, error mark options={black, line width=0.7pt, mark size=2.5pt, rotate=90}] coordinates {(M,35.280) += (0,0.714) -= (0,0.721)};
\nextgroupplot[title={Parler-mini}]
\addplot[only marks, color=palLightBlue, mark=o, mark size=1.6pt, error bars/.cd, y dir=both, y explicit, error bar style={line width=0.7pt, black}, error mark options={black, line width=0.7pt, mark size=2.5pt, rotate=90}] coordinates {(F,30.590) += (0,0.379) -= (0,0.373)};
\addplot[only marks, color=palLightBlue, mark=*, mark size=1.6pt, error bars/.cd, y dir=both, y explicit, error bar style={line width=0.7pt, black}, error mark options={black, line width=0.7pt, mark size=2.5pt, rotate=90}] coordinates {(M,35.036) += (0,0.484) -= (0,0.490)};
\nextgroupplot[title={CapSpeech-NAR}]
\addplot[only marks, color=palGreen, mark=o, mark size=1.6pt, error bars/.cd, y dir=both, y explicit, error bar style={line width=0.7pt, black}, error mark options={black, line width=0.7pt, mark size=2.5pt, rotate=90}] coordinates {(F,17.206) += (0,0.313) -= (0,0.318)};
\addplot[only marks, color=palGreen, mark=*, mark size=1.6pt, error bars/.cd, y dir=both, y explicit, error bar style={line width=0.7pt, black}, error mark options={black, line width=0.7pt, mark size=2.5pt, rotate=90}] coordinates {(M,20.688) += (0,0.393) -= (0,0.396)};
\nextgroupplot[title={Qwen3-TTS}]
\addplot[only marks, color=palOrange, mark=o, mark size=1.6pt, error bars/.cd, y dir=both, y explicit, error bar style={line width=0.7pt, black}, error mark options={black, line width=0.7pt, mark size=2.5pt, rotate=90}] coordinates {(F,16.552) += (0,0.342) -= (0,0.343)};
\addplot[only marks, color=palOrange, mark=*, mark size=1.6pt, error bars/.cd, y dir=both, y explicit, error bar style={line width=0.7pt, black}, error mark options={black, line width=0.7pt, mark size=2.5pt, rotate=90}] coordinates {(M,28.078) += (0,0.500) -= (0,0.498)};
\nextgroupplot[title={VoxInstruct}]
\addplot[only marks, color=palDarkPurple, mark=o, mark size=1.6pt, error bars/.cd, y dir=both, y explicit, error bar style={line width=0.7pt, black}, error mark options={black, line width=0.7pt, mark size=2.5pt, rotate=90}] coordinates {(F,16.175) += (0,0.404) -= (0,0.402)};
\addplot[only marks, color=palDarkPurple, mark=*, mark size=1.6pt, error bars/.cd, y dir=both, y explicit, error bar style={line width=0.7pt, black}, error mark options={black, line width=0.7pt, mark size=2.5pt, rotate=90}] coordinates {(M,24.624) += (0,0.501) -= (0,0.502)};
\nextgroupplot[title={VoxCPM2}]
\addplot[only marks, color=palMidPurple, mark=o, mark size=1.6pt, error bars/.cd, y dir=both, y explicit, error bar style={line width=0.7pt, black}, error mark options={black, line width=0.7pt, mark size=2.5pt, rotate=90}] coordinates {(F,9.263) += (0,0.244) -= (0,0.237)};
\addplot[only marks, color=palMidPurple, mark=*, mark size=1.6pt, error bars/.cd, y dir=both, y explicit, error bar style={line width=0.7pt, black}, error mark options={black, line width=0.7pt, mark size=2.5pt, rotate=90}] coordinates {(M,7.983) += (0,0.182) -= (0,0.180)};
\end{groupplot}\end{tikzpicture}
 
\vspace{2pt}
{\small\textbf{(b) Age}}\\[1pt]
\begin{tikzpicture}\begin{groupplot}[group style={group size=7 by 1, horizontal sep=0.68cm}, scale only axis, width=1.55cm, height=2.6cm, symbolic x coords={C,Y,Mi,E}, xtick={C,Y,Mi,E}, enlarge x limits=0.15, tick label style={font=\tiny}, x tick label style={rotate=90, anchor=east}, title style={font=\scriptsize}, ymajorgrids, grid style={draw=gray!20}, yticklabel style={/pgf/number format/fixed, /pgf/number format/precision=1}]
\nextgroupplot[title={MOSS-VG}, ylabel={Vendi}, ylabel style={font=\scriptsize}]
\addplot[only marks, color=palRed, mark=o, mark size=1.6pt, error bars/.cd, y dir=both, y explicit, error bar style={line width=0.7pt, black}, error mark options={black, line width=0.7pt, mark size=2.5pt, rotate=90}] coordinates {(C,25.058) += (0,0.584) -= (0,0.598)};
\addplot[only marks, color=palRed, mark=square*, mark size=1.6pt, error bars/.cd, y dir=both, y explicit, error bar style={line width=0.7pt, black}, error mark options={black, line width=0.7pt, mark size=2.5pt, rotate=90}] coordinates {(Y,55.502) += (0,0.640) -= (0,0.643)};
\addplot[only marks, color=palRed, mark=triangle*, mark size=1.6pt, error bars/.cd, y dir=both, y explicit, error bar style={line width=0.7pt, black}, error mark options={black, line width=0.7pt, mark size=2.5pt, rotate=90}] coordinates {(Mi,57.536) += (0,0.645) -= (0,0.676)};
\addplot[only marks, color=palRed, mark=diamond*, mark size=1.6pt, error bars/.cd, y dir=both, y explicit, error bar style={line width=0.7pt, black}, error mark options={black, line width=0.7pt, mark size=2.5pt, rotate=90}] coordinates {(E,57.759) += (0,0.620) -= (0,0.615)};
\nextgroupplot[title={Parler-large}]
\addplot[only marks, color=palDarkBlue, mark=o, mark size=1.6pt, error bars/.cd, y dir=both, y explicit, error bar style={line width=0.7pt, black}, error mark options={black, line width=0.7pt, mark size=2.5pt, rotate=90}] coordinates {(C,46.718) += (0,0.532) -= (0,0.526)};
\addplot[only marks, color=palDarkBlue, mark=square*, mark size=1.6pt, error bars/.cd, y dir=both, y explicit, error bar style={line width=0.7pt, black}, error mark options={black, line width=0.7pt, mark size=2.5pt, rotate=90}] coordinates {(Y,46.307) += (0,0.575) -= (0,0.571)};
\addplot[only marks, color=palDarkBlue, mark=triangle*, mark size=1.6pt, error bars/.cd, y dir=both, y explicit, error bar style={line width=0.7pt, black}, error mark options={black, line width=0.7pt, mark size=2.5pt, rotate=90}] coordinates {(Mi,48.178) += (0,0.584) -= (0,0.590)};
\addplot[only marks, color=palDarkBlue, mark=diamond*, mark size=1.6pt, error bars/.cd, y dir=both, y explicit, error bar style={line width=0.7pt, black}, error mark options={black, line width=0.7pt, mark size=2.5pt, rotate=90}] coordinates {(E,49.010) += (0,0.495) -= (0,0.491)};
\nextgroupplot[title={Parler-mini}]
\addplot[only marks, color=palLightBlue, mark=o, mark size=1.6pt, error bars/.cd, y dir=both, y explicit, error bar style={line width=0.7pt, black}, error mark options={black, line width=0.7pt, mark size=2.5pt, rotate=90}] coordinates {(C,39.467) += (0,0.496) -= (0,0.493)};
\addplot[only marks, color=palLightBlue, mark=square*, mark size=1.6pt, error bars/.cd, y dir=both, y explicit, error bar style={line width=0.7pt, black}, error mark options={black, line width=0.7pt, mark size=2.5pt, rotate=90}] coordinates {(Y,41.132) += (0,0.496) -= (0,0.491)};
\addplot[only marks, color=palLightBlue, mark=triangle*, mark size=1.6pt, error bars/.cd, y dir=both, y explicit, error bar style={line width=0.7pt, black}, error mark options={black, line width=0.7pt, mark size=2.5pt, rotate=90}] coordinates {(Mi,41.197) += (0,0.473) -= (0,0.481)};
\addplot[only marks, color=palLightBlue, mark=diamond*, mark size=1.6pt, error bars/.cd, y dir=both, y explicit, error bar style={line width=0.7pt, black}, error mark options={black, line width=0.7pt, mark size=2.5pt, rotate=90}] coordinates {(E,40.229) += (0,0.528) -= (0,0.527)};
\nextgroupplot[title={CapSpeech-NAR}]
\addplot[only marks, color=palGreen, mark=o, mark size=1.6pt, error bars/.cd, y dir=both, y explicit, error bar style={line width=0.7pt, black}, error mark options={black, line width=0.7pt, mark size=2.5pt, rotate=90}] coordinates {(C,23.903) += (0,0.496) -= (0,0.496)};
\addplot[only marks, color=palGreen, mark=square*, mark size=1.6pt, error bars/.cd, y dir=both, y explicit, error bar style={line width=0.7pt, black}, error mark options={black, line width=0.7pt, mark size=2.5pt, rotate=90}] coordinates {(Y,16.964) += (0,0.425) -= (0,0.421)};
\addplot[only marks, color=palGreen, mark=triangle*, mark size=1.6pt, error bars/.cd, y dir=both, y explicit, error bar style={line width=0.7pt, black}, error mark options={black, line width=0.7pt, mark size=2.5pt, rotate=90}] coordinates {(Mi,27.364) += (0,0.556) -= (0,0.547)};
\addplot[only marks, color=palGreen, mark=diamond*, mark size=1.6pt, error bars/.cd, y dir=both, y explicit, error bar style={line width=0.7pt, black}, error mark options={black, line width=0.7pt, mark size=2.5pt, rotate=90}] coordinates {(E,26.859) += (0,0.516) -= (0,0.506)};
\nextgroupplot[title={Qwen3-TTS}]
\addplot[only marks, color=palOrange, mark=o, mark size=1.6pt, error bars/.cd, y dir=both, y explicit, error bar style={line width=0.7pt, black}, error mark options={black, line width=0.7pt, mark size=2.5pt, rotate=90}] coordinates {(C,5.515) += (0,0.108) -= (0,0.109)};
\addplot[only marks, color=palOrange, mark=square*, mark size=1.6pt, error bars/.cd, y dir=both, y explicit, error bar style={line width=0.7pt, black}, error mark options={black, line width=0.7pt, mark size=2.5pt, rotate=90}] coordinates {(Y,15.275) += (0,0.469) -= (0,0.454)};
\addplot[only marks, color=palOrange, mark=triangle*, mark size=1.6pt, error bars/.cd, y dir=both, y explicit, error bar style={line width=0.7pt, black}, error mark options={black, line width=0.7pt, mark size=2.5pt, rotate=90}] coordinates {(Mi,31.394) += (0,0.615) -= (0,0.628)};
\addplot[only marks, color=palOrange, mark=diamond*, mark size=1.6pt, error bars/.cd, y dir=both, y explicit, error bar style={line width=0.7pt, black}, error mark options={black, line width=0.7pt, mark size=2.5pt, rotate=90}] coordinates {(E,33.632) += (0,0.699) -= (0,0.690)};
\nextgroupplot[title={VoxInstruct}]
\addplot[only marks, color=palDarkPurple, mark=o, mark size=1.6pt, error bars/.cd, y dir=both, y explicit, error bar style={line width=0.7pt, black}, error mark options={black, line width=0.7pt, mark size=2.5pt, rotate=90}] coordinates {(C,19.701) += (0,0.537) -= (0,0.526)};
\addplot[only marks, color=palDarkPurple, mark=square*, mark size=1.6pt, error bars/.cd, y dir=both, y explicit, error bar style={line width=0.7pt, black}, error mark options={black, line width=0.7pt, mark size=2.5pt, rotate=90}] coordinates {(Y,22.807) += (0,0.487) -= (0,0.490)};
\addplot[only marks, color=palDarkPurple, mark=triangle*, mark size=1.6pt, error bars/.cd, y dir=both, y explicit, error bar style={line width=0.7pt, black}, error mark options={black, line width=0.7pt, mark size=2.5pt, rotate=90}] coordinates {(Mi,25.479) += (0,0.523) -= (0,0.522)};
\addplot[only marks, color=palDarkPurple, mark=diamond*, mark size=1.6pt, error bars/.cd, y dir=both, y explicit, error bar style={line width=0.7pt, black}, error mark options={black, line width=0.7pt, mark size=2.5pt, rotate=90}] coordinates {(E,28.384) += (0,0.537) -= (0,0.551)};
\nextgroupplot[title={VoxCPM2}]
\addplot[only marks, color=palMidPurple, mark=o, mark size=1.6pt, error bars/.cd, y dir=both, y explicit, error bar style={line width=0.7pt, black}, error mark options={black, line width=0.7pt, mark size=2.5pt, rotate=90}] coordinates {(C,23.708) += (0,0.564) -= (0,0.560)};
\addplot[only marks, color=palMidPurple, mark=square*, mark size=1.6pt, error bars/.cd, y dir=both, y explicit, error bar style={line width=0.7pt, black}, error mark options={black, line width=0.7pt, mark size=2.5pt, rotate=90}] coordinates {(Y,15.575) += (0,0.457) -= (0,0.439)};
\addplot[only marks, color=palMidPurple, mark=triangle*, mark size=1.6pt, error bars/.cd, y dir=both, y explicit, error bar style={line width=0.7pt, black}, error mark options={black, line width=0.7pt, mark size=2.5pt, rotate=90}] coordinates {(Mi,12.835) += (0,0.355) -= (0,0.352)};
\addplot[only marks, color=palMidPurple, mark=diamond*, mark size=1.6pt, error bars/.cd, y dir=both, y explicit, error bar style={line width=0.7pt, black}, error mark options={black, line width=0.7pt, mark size=2.5pt, rotate=90}] coordinates {(E,16.303) += (0,0.422) -= (0,0.417)};
\end{groupplot}\end{tikzpicture}
 
\vspace{2pt}
{\small\textbf{(c) Accent}}\\[1pt]
\begin{tikzpicture}\begin{groupplot}[group style={group size=7 by 1, horizontal sep=0.68cm}, scale only axis, width=1.55cm, height=2.6cm, symbolic x coords={NA,Br,Oc,SA,SAf,Ge}, xtick={NA,Br,Oc,SA,SAf,Ge}, enlarge x limits=0.15, tick label style={font=\tiny}, x tick label style={rotate=90, anchor=east}, title style={font=\scriptsize}, ymajorgrids, grid style={draw=gray!20}, yticklabel style={/pgf/number format/fixed, /pgf/number format/precision=1}]
\nextgroupplot[title={MOSS-VG}, ylabel={Vendi}, ylabel style={font=\scriptsize}]
\addplot[only marks, color=palRed, mark=o, mark size=1.6pt, error bars/.cd, y dir=both, y explicit, error bar style={line width=0.7pt, black}, error mark options={black, line width=0.7pt, mark size=2.5pt, rotate=90}] coordinates {(NA,60.300) += (0,0.629) -= (0,0.643)};
\addplot[only marks, color=palRed, mark=square*, mark size=1.6pt, error bars/.cd, y dir=both, y explicit, error bar style={line width=0.7pt, black}, error mark options={black, line width=0.7pt, mark size=2.5pt, rotate=90}] coordinates {(Br,54.002) += (0,0.636) -= (0,0.629)};
\addplot[only marks, color=palRed, mark=triangle*, mark size=1.6pt, error bars/.cd, y dir=both, y explicit, error bar style={line width=0.7pt, black}, error mark options={black, line width=0.7pt, mark size=2.5pt, rotate=90}] coordinates {(Oc,56.240) += (0,0.694) -= (0,0.713)};
\addplot[only marks, color=palRed, mark=diamond*, mark size=1.6pt, error bars/.cd, y dir=both, y explicit, error bar style={line width=0.7pt, black}, error mark options={black, line width=0.7pt, mark size=2.5pt, rotate=90}] coordinates {(SA,60.159) += (0,0.680) -= (0,0.691)};
\addplot[only marks, color=palRed, mark=pentagon*, mark size=1.6pt, error bars/.cd, y dir=both, y explicit, error bar style={line width=0.7pt, black}, error mark options={black, line width=0.7pt, mark size=2.5pt, rotate=90}] coordinates {(SAf,56.470) += (0,0.600) -= (0,0.616)};
\addplot[only marks, color=palRed, mark=star, mark size=2.2pt, error bars/.cd, y dir=both, y explicit, error bar style={line width=0.7pt, black}, error mark options={black, line width=0.7pt, mark size=2.5pt, rotate=90}] coordinates {(Ge,56.565) += (0,0.633) -= (0,0.643)};
\nextgroupplot[title={Parler-large}]
\addplot[only marks, color=palDarkBlue, mark=o, mark size=1.6pt, error bars/.cd, y dir=both, y explicit, error bar style={line width=0.7pt, black}, error mark options={black, line width=0.7pt, mark size=2.5pt, rotate=90}] coordinates {(NA,49.246) += (0,0.523) -= (0,0.520)};
\addplot[only marks, color=palDarkBlue, mark=square*, mark size=1.6pt, error bars/.cd, y dir=both, y explicit, error bar style={line width=0.7pt, black}, error mark options={black, line width=0.7pt, mark size=2.5pt, rotate=90}] coordinates {(Br,48.554) += (0,0.522) -= (0,0.524)};
\addplot[only marks, color=palDarkBlue, mark=triangle*, mark size=1.6pt, error bars/.cd, y dir=both, y explicit, error bar style={line width=0.7pt, black}, error mark options={black, line width=0.7pt, mark size=2.5pt, rotate=90}] coordinates {(Oc,48.892) += (0,0.501) -= (0,0.491)};
\addplot[only marks, color=palDarkBlue, mark=diamond*, mark size=1.6pt, error bars/.cd, y dir=both, y explicit, error bar style={line width=0.7pt, black}, error mark options={black, line width=0.7pt, mark size=2.5pt, rotate=90}] coordinates {(SA,47.199) += (0,0.530) -= (0,0.527)};
\addplot[only marks, color=palDarkBlue, mark=pentagon*, mark size=1.6pt, error bars/.cd, y dir=both, y explicit, error bar style={line width=0.7pt, black}, error mark options={black, line width=0.7pt, mark size=2.5pt, rotate=90}] coordinates {(SAf,45.719) += (0,0.556) -= (0,0.564)};
\addplot[only marks, color=palDarkBlue, mark=star, mark size=2.2pt, error bars/.cd, y dir=both, y explicit, error bar style={line width=0.7pt, black}, error mark options={black, line width=0.7pt, mark size=2.5pt, rotate=90}] coordinates {(Ge,46.689) += (0,0.611) -= (0,0.617)};
\nextgroupplot[title={Parler-mini}]
\addplot[only marks, color=palLightBlue, mark=o, mark size=1.6pt, error bars/.cd, y dir=both, y explicit, error bar style={line width=0.7pt, black}, error mark options={black, line width=0.7pt, mark size=2.5pt, rotate=90}] coordinates {(NA,41.607) += (0,0.477) -= (0,0.474)};
\addplot[only marks, color=palLightBlue, mark=square*, mark size=1.6pt, error bars/.cd, y dir=both, y explicit, error bar style={line width=0.7pt, black}, error mark options={black, line width=0.7pt, mark size=2.5pt, rotate=90}] coordinates {(Br,41.532) += (0,0.475) -= (0,0.476)};
\addplot[only marks, color=palLightBlue, mark=triangle*, mark size=1.6pt, error bars/.cd, y dir=both, y explicit, error bar style={line width=0.7pt, black}, error mark options={black, line width=0.7pt, mark size=2.5pt, rotate=90}] coordinates {(Oc,39.496) += (0,0.498) -= (0,0.511)};
\addplot[only marks, color=palLightBlue, mark=diamond*, mark size=1.6pt, error bars/.cd, y dir=both, y explicit, error bar style={line width=0.7pt, black}, error mark options={black, line width=0.7pt, mark size=2.5pt, rotate=90}] coordinates {(SA,39.602) += (0,0.507) -= (0,0.506)};
\addplot[only marks, color=palLightBlue, mark=pentagon*, mark size=1.6pt, error bars/.cd, y dir=both, y explicit, error bar style={line width=0.7pt, black}, error mark options={black, line width=0.7pt, mark size=2.5pt, rotate=90}] coordinates {(SAf,38.879) += (0,0.505) -= (0,0.498)};
\addplot[only marks, color=palLightBlue, mark=star, mark size=2.2pt, error bars/.cd, y dir=both, y explicit, error bar style={line width=0.7pt, black}, error mark options={black, line width=0.7pt, mark size=2.5pt, rotate=90}] coordinates {(Ge,40.959) += (0,0.462) -= (0,0.460)};
\nextgroupplot[title={CapSpeech-NAR}]
\addplot[only marks, color=palGreen, mark=o, mark size=1.6pt, error bars/.cd, y dir=both, y explicit, error bar style={line width=0.7pt, black}, error mark options={black, line width=0.7pt, mark size=2.5pt, rotate=90}] coordinates {(NA,15.885) += (0,0.248) -= (0,0.246)};
\addplot[only marks, color=palGreen, mark=square*, mark size=1.6pt, error bars/.cd, y dir=both, y explicit, error bar style={line width=0.7pt, black}, error mark options={black, line width=0.7pt, mark size=2.5pt, rotate=90}] coordinates {(Br,16.306) += (0,0.281) -= (0,0.276)};
\addplot[only marks, color=palGreen, mark=triangle*, mark size=1.6pt, error bars/.cd, y dir=both, y explicit, error bar style={line width=0.7pt, black}, error mark options={black, line width=0.7pt, mark size=2.5pt, rotate=90}] coordinates {(Oc,9.783) += (0,0.200) -= (0,0.198)};
\addplot[only marks, color=palGreen, mark=diamond*, mark size=1.6pt, error bars/.cd, y dir=both, y explicit, error bar style={line width=0.7pt, black}, error mark options={black, line width=0.7pt, mark size=2.5pt, rotate=90}] coordinates {(SA,6.608) += (0,0.158) -= (0,0.157)};
\addplot[only marks, color=palGreen, mark=pentagon*, mark size=1.6pt, error bars/.cd, y dir=both, y explicit, error bar style={line width=0.7pt, black}, error mark options={black, line width=0.7pt, mark size=2.5pt, rotate=90}] coordinates {(SAf,8.150) += (0,0.157) -= (0,0.155)};
\addplot[only marks, color=palGreen, mark=star, mark size=2.2pt, error bars/.cd, y dir=both, y explicit, error bar style={line width=0.7pt, black}, error mark options={black, line width=0.7pt, mark size=2.5pt, rotate=90}] coordinates {(Ge,15.928) += (0,0.325) -= (0,0.319)};
\nextgroupplot[title={Qwen3-TTS}]
\addplot[only marks, color=palOrange, mark=o, mark size=1.6pt, error bars/.cd, y dir=both, y explicit, error bar style={line width=0.7pt, black}, error mark options={black, line width=0.7pt, mark size=2.5pt, rotate=90}] coordinates {(NA,29.564) += (0,0.550) -= (0,0.567)};
\addplot[only marks, color=palOrange, mark=square*, mark size=1.6pt, error bars/.cd, y dir=both, y explicit, error bar style={line width=0.7pt, black}, error mark options={black, line width=0.7pt, mark size=2.5pt, rotate=90}] coordinates {(Br,31.081) += (0,0.597) -= (0,0.608)};
\addplot[only marks, color=palOrange, mark=triangle*, mark size=1.6pt, error bars/.cd, y dir=both, y explicit, error bar style={line width=0.7pt, black}, error mark options={black, line width=0.7pt, mark size=2.5pt, rotate=90}] coordinates {(Oc,28.672) += (0,0.639) -= (0,0.639)};
\addplot[only marks, color=palOrange, mark=diamond*, mark size=1.6pt, error bars/.cd, y dir=both, y explicit, error bar style={line width=0.7pt, black}, error mark options={black, line width=0.7pt, mark size=2.5pt, rotate=90}] coordinates {(SA,28.202) += (0,0.792) -= (0,0.792)};
\addplot[only marks, color=palOrange, mark=pentagon*, mark size=1.6pt, error bars/.cd, y dir=both, y explicit, error bar style={line width=0.7pt, black}, error mark options={black, line width=0.7pt, mark size=2.5pt, rotate=90}] coordinates {(SAf,29.587) += (0,0.770) -= (0,0.782)};
\addplot[only marks, color=palOrange, mark=star, mark size=2.2pt, error bars/.cd, y dir=both, y explicit, error bar style={line width=0.7pt, black}, error mark options={black, line width=0.7pt, mark size=2.5pt, rotate=90}] coordinates {(Ge,32.659) += (0,0.612) -= (0,0.625)};
\nextgroupplot[title={VoxInstruct}]
\addplot[only marks, color=palDarkPurple, mark=o, mark size=1.6pt, error bars/.cd, y dir=both, y explicit, error bar style={line width=0.7pt, black}, error mark options={black, line width=0.7pt, mark size=2.5pt, rotate=90}] coordinates {(NA,21.781) += (0,0.420) -= (0,0.426)};
\addplot[only marks, color=palDarkPurple, mark=square*, mark size=1.6pt, error bars/.cd, y dir=both, y explicit, error bar style={line width=0.7pt, black}, error mark options={black, line width=0.7pt, mark size=2.5pt, rotate=90}] coordinates {(Br,21.053) += (0,0.431) -= (0,0.428)};
\addplot[only marks, color=palDarkPurple, mark=triangle*, mark size=1.6pt, error bars/.cd, y dir=both, y explicit, error bar style={line width=0.7pt, black}, error mark options={black, line width=0.7pt, mark size=2.5pt, rotate=90}] coordinates {(Oc,21.230) += (0,0.430) -= (0,0.434)};
\addplot[only marks, color=palDarkPurple, mark=diamond*, mark size=1.6pt, error bars/.cd, y dir=both, y explicit, error bar style={line width=0.7pt, black}, error mark options={black, line width=0.7pt, mark size=2.5pt, rotate=90}] coordinates {(SA,21.375) += (0,0.472) -= (0,0.474)};
\addplot[only marks, color=palDarkPurple, mark=pentagon*, mark size=1.6pt, error bars/.cd, y dir=both, y explicit, error bar style={line width=0.7pt, black}, error mark options={black, line width=0.7pt, mark size=2.5pt, rotate=90}] coordinates {(SAf,21.669) += (0,0.439) -= (0,0.440)};
\addplot[only marks, color=palDarkPurple, mark=star, mark size=2.2pt, error bars/.cd, y dir=both, y explicit, error bar style={line width=0.7pt, black}, error mark options={black, line width=0.7pt, mark size=2.5pt, rotate=90}] coordinates {(Ge,21.078) += (0,0.438) -= (0,0.435)};
\nextgroupplot[title={VoxCPM2}]
\addplot[only marks, color=palMidPurple, mark=o, mark size=1.6pt, error bars/.cd, y dir=both, y explicit, error bar style={line width=0.7pt, black}, error mark options={black, line width=0.7pt, mark size=2.5pt, rotate=90}] coordinates {(NA,12.759) += (0,0.302) -= (0,0.296)};
\addplot[only marks, color=palMidPurple, mark=square*, mark size=1.6pt, error bars/.cd, y dir=both, y explicit, error bar style={line width=0.7pt, black}, error mark options={black, line width=0.7pt, mark size=2.5pt, rotate=90}] coordinates {(Br,15.684) += (0,0.364) -= (0,0.356)};
\addplot[only marks, color=palMidPurple, mark=triangle*, mark size=1.6pt, error bars/.cd, y dir=both, y explicit, error bar style={line width=0.7pt, black}, error mark options={black, line width=0.7pt, mark size=2.5pt, rotate=90}] coordinates {(Oc,15.426) += (0,0.349) -= (0,0.354)};
\addplot[only marks, color=palMidPurple, mark=diamond*, mark size=1.6pt, error bars/.cd, y dir=both, y explicit, error bar style={line width=0.7pt, black}, error mark options={black, line width=0.7pt, mark size=2.5pt, rotate=90}] coordinates {(SA,14.842) += (0,0.368) -= (0,0.364)};
\addplot[only marks, color=palMidPurple, mark=pentagon*, mark size=1.6pt, error bars/.cd, y dir=both, y explicit, error bar style={line width=0.7pt, black}, error mark options={black, line width=0.7pt, mark size=2.5pt, rotate=90}] coordinates {(SAf,17.160) += (0,0.404) -= (0,0.405)};
\addplot[only marks, color=palMidPurple, mark=star, mark size=2.2pt, error bars/.cd, y dir=both, y explicit, error bar style={line width=0.7pt, black}, error mark options={black, line width=0.7pt, mark size=2.5pt, rotate=90}] coordinates {(Ge,14.353) += (0,0.323) -= (0,0.314)};
\end{groupplot}\end{tikzpicture}
 
\vspace{2pt}
{\small\textbf{(d) Emotion}}\\[1pt]
\begin{tikzpicture}\begin{groupplot}[group style={group size=7 by 1, horizontal sep=0.68cm}, scale only axis, width=1.55cm, height=2.6cm, symbolic x coords={Ang,Hap,Neu,Sad}, xtick={Ang,Hap,Neu,Sad}, enlarge x limits=0.15, tick label style={font=\tiny}, x tick label style={rotate=90, anchor=east}, title style={font=\scriptsize}, ymajorgrids, grid style={draw=gray!20}, yticklabel style={/pgf/number format/fixed, /pgf/number format/precision=1}]
\nextgroupplot[title={MOSS-VG}, ylabel={Vendi}, ylabel style={font=\scriptsize}]
\addplot[only marks, color=palRed, mark=o, mark size=1.6pt, error bars/.cd, y dir=both, y explicit, error bar style={line width=0.7pt, black}, error mark options={black, line width=0.7pt, mark size=2.5pt, rotate=90}] coordinates {(Ang,59.137) += (0,0.622) -= (0,0.639)};
\addplot[only marks, color=palRed, mark=square*, mark size=1.6pt, error bars/.cd, y dir=both, y explicit, error bar style={line width=0.7pt, black}, error mark options={black, line width=0.7pt, mark size=2.5pt, rotate=90}] coordinates {(Hap,58.258) += (0,0.750) -= (0,0.760)};
\addplot[only marks, color=palRed, mark=triangle*, mark size=1.6pt, error bars/.cd, y dir=both, y explicit, error bar style={line width=0.7pt, black}, error mark options={black, line width=0.7pt, mark size=2.5pt, rotate=90}] coordinates {(Neu,58.355) += (0,0.633) -= (0,0.628)};
\addplot[only marks, color=palRed, mark=diamond*, mark size=1.6pt, error bars/.cd, y dir=both, y explicit, error bar style={line width=0.7pt, black}, error mark options={black, line width=0.7pt, mark size=2.5pt, rotate=90}] coordinates {(Sad,57.635) += (0,0.875) -= (0,0.884)};
\nextgroupplot[title={Parler-large}]
\addplot[only marks, color=palDarkBlue, mark=o, mark size=1.6pt, error bars/.cd, y dir=both, y explicit, error bar style={line width=0.7pt, black}, error mark options={black, line width=0.7pt, mark size=2.5pt, rotate=90}] coordinates {(Ang,46.998) += (0,0.515) -= (0,0.506)};
\addplot[only marks, color=palDarkBlue, mark=square*, mark size=1.6pt, error bars/.cd, y dir=both, y explicit, error bar style={line width=0.7pt, black}, error mark options={black, line width=0.7pt, mark size=2.5pt, rotate=90}] coordinates {(Hap,47.065) += (0,0.503) -= (0,0.506)};
\addplot[only marks, color=palDarkBlue, mark=triangle*, mark size=1.6pt, error bars/.cd, y dir=both, y explicit, error bar style={line width=0.7pt, black}, error mark options={black, line width=0.7pt, mark size=2.5pt, rotate=90}] coordinates {(Neu,48.888) += (0,0.538) -= (0,0.545)};
\addplot[only marks, color=palDarkBlue, mark=diamond*, mark size=1.6pt, error bars/.cd, y dir=both, y explicit, error bar style={line width=0.7pt, black}, error mark options={black, line width=0.7pt, mark size=2.5pt, rotate=90}] coordinates {(Sad,47.799) += (0,0.479) -= (0,0.485)};
\nextgroupplot[title={Parler-mini}]
\addplot[only marks, color=palLightBlue, mark=o, mark size=1.6pt, error bars/.cd, y dir=both, y explicit, error bar style={line width=0.7pt, black}, error mark options={black, line width=0.7pt, mark size=2.5pt, rotate=90}] coordinates {(Ang,39.664) += (0,0.483) -= (0,0.489)};
\addplot[only marks, color=palLightBlue, mark=square*, mark size=1.6pt, error bars/.cd, y dir=both, y explicit, error bar style={line width=0.7pt, black}, error mark options={black, line width=0.7pt, mark size=2.5pt, rotate=90}] coordinates {(Hap,39.046) += (0,0.479) -= (0,0.494)};
\addplot[only marks, color=palLightBlue, mark=triangle*, mark size=1.6pt, error bars/.cd, y dir=both, y explicit, error bar style={line width=0.7pt, black}, error mark options={black, line width=0.7pt, mark size=2.5pt, rotate=90}] coordinates {(Neu,40.732) += (0,0.469) -= (0,0.473)};
\addplot[only marks, color=palLightBlue, mark=diamond*, mark size=1.6pt, error bars/.cd, y dir=both, y explicit, error bar style={line width=0.7pt, black}, error mark options={black, line width=0.7pt, mark size=2.5pt, rotate=90}] coordinates {(Sad,39.230) += (0,0.472) -= (0,0.467)};
\nextgroupplot[title={CapSpeech-NAR}]
\addplot[only marks, color=palGreen, mark=o, mark size=1.6pt, error bars/.cd, y dir=both, y explicit, error bar style={line width=0.7pt, black}, error mark options={black, line width=0.7pt, mark size=2.5pt, rotate=90}] coordinates {(Ang,7.698) += (0,0.201) -= (0,0.194)};
\addplot[only marks, color=palGreen, mark=square*, mark size=1.6pt, error bars/.cd, y dir=both, y explicit, error bar style={line width=0.7pt, black}, error mark options={black, line width=0.7pt, mark size=2.5pt, rotate=90}] coordinates {(Hap,10.499) += (0,0.281) -= (0,0.276)};
\addplot[only marks, color=palGreen, mark=triangle*, mark size=1.6pt, error bars/.cd, y dir=both, y explicit, error bar style={line width=0.7pt, black}, error mark options={black, line width=0.7pt, mark size=2.5pt, rotate=90}] coordinates {(Neu,10.272) += (0,0.307) -= (0,0.308)};
\addplot[only marks, color=palGreen, mark=diamond*, mark size=1.6pt, error bars/.cd, y dir=both, y explicit, error bar style={line width=0.7pt, black}, error mark options={black, line width=0.7pt, mark size=2.5pt, rotate=90}] coordinates {(Sad,10.372) += (0,0.311) -= (0,0.312)};
\nextgroupplot[title={Qwen3-TTS}]
\addplot[only marks, color=palOrange, mark=o, mark size=1.6pt, error bars/.cd, y dir=both, y explicit, error bar style={line width=0.7pt, black}, error mark options={black, line width=0.7pt, mark size=2.5pt, rotate=90}] coordinates {(Ang,28.016) += (0,0.481) -= (0,0.493)};
\addplot[only marks, color=palOrange, mark=square*, mark size=1.6pt, error bars/.cd, y dir=both, y explicit, error bar style={line width=0.7pt, black}, error mark options={black, line width=0.7pt, mark size=2.5pt, rotate=90}] coordinates {(Hap,21.931) += (0,0.522) -= (0,0.516)};
\addplot[only marks, color=palOrange, mark=triangle*, mark size=1.6pt, error bars/.cd, y dir=both, y explicit, error bar style={line width=0.7pt, black}, error mark options={black, line width=0.7pt, mark size=2.5pt, rotate=90}] coordinates {(Neu,31.981) += (0,0.478) -= (0,0.481)};
\addplot[only marks, color=palOrange, mark=diamond*, mark size=1.6pt, error bars/.cd, y dir=both, y explicit, error bar style={line width=0.7pt, black}, error mark options={black, line width=0.7pt, mark size=2.5pt, rotate=90}] coordinates {(Sad,28.471) += (0,0.848) -= (0,0.820)};
\nextgroupplot[title={VoxInstruct}]
\addplot[only marks, color=palDarkPurple, mark=o, mark size=1.6pt, error bars/.cd, y dir=both, y explicit, error bar style={line width=0.7pt, black}, error mark options={black, line width=0.7pt, mark size=2.5pt, rotate=90}] coordinates {(Ang,24.691) += (0,0.510) -= (0,0.510)};
\addplot[only marks, color=palDarkPurple, mark=square*, mark size=1.6pt, error bars/.cd, y dir=both, y explicit, error bar style={line width=0.7pt, black}, error mark options={black, line width=0.7pt, mark size=2.5pt, rotate=90}] coordinates {(Hap,21.673) += (0,0.491) -= (0,0.486)};
\addplot[only marks, color=palDarkPurple, mark=triangle*, mark size=1.6pt, error bars/.cd, y dir=both, y explicit, error bar style={line width=0.7pt, black}, error mark options={black, line width=0.7pt, mark size=2.5pt, rotate=90}] coordinates {(Neu,23.057) += (0,0.502) -= (0,0.502)};
\addplot[only marks, color=palDarkPurple, mark=diamond*, mark size=1.6pt, error bars/.cd, y dir=both, y explicit, error bar style={line width=0.7pt, black}, error mark options={black, line width=0.7pt, mark size=2.5pt, rotate=90}] coordinates {(Sad,23.091) += (0,0.518) -= (0,0.530)};
\nextgroupplot[title={VoxCPM2}]
\addplot[only marks, color=palMidPurple, mark=o, mark size=1.6pt, error bars/.cd, y dir=both, y explicit, error bar style={line width=0.7pt, black}, error mark options={black, line width=0.7pt, mark size=2.5pt, rotate=90}] coordinates {(Ang,10.883) += (0,0.223) -= (0,0.212)};
\addplot[only marks, color=palMidPurple, mark=square*, mark size=1.6pt, error bars/.cd, y dir=both, y explicit, error bar style={line width=0.7pt, black}, error mark options={black, line width=0.7pt, mark size=2.5pt, rotate=90}] coordinates {(Hap,11.303) += (0,0.245) -= (0,0.242)};
\addplot[only marks, color=palMidPurple, mark=triangle*, mark size=1.6pt, error bars/.cd, y dir=both, y explicit, error bar style={line width=0.7pt, black}, error mark options={black, line width=0.7pt, mark size=2.5pt, rotate=90}] coordinates {(Neu,9.984) += (0,0.189) -= (0,0.181)};
\addplot[only marks, color=palMidPurple, mark=diamond*, mark size=1.6pt, error bars/.cd, y dir=both, y explicit, error bar style={line width=0.7pt, black}, error mark options={black, line width=0.7pt, mark size=2.5pt, rotate=90}] coordinates {(Sad,10.850) += (0,0.218) -= (0,0.210)};
\end{groupplot}\end{tikzpicture}
\caption{\textbf{Speaker diversity with 95\% confidence intervals for each system.} Vendi of each prompted attribute for (a) gender, (b) age, (c) accent, and (d) emotion, with black bars showing 95\% confidence intervals. Each panel has its own y-axis. Abbreviations are F female, M male, C child, Y young adult, Mi middle-aged, E elderly, NA North American, Br British, Oc Oceanian, SA South Asian, SAf South African, Ge Germanic, Ang angry, Hap happy, Neu neutral, and Sad sad.}
\label{fig:per-system}
\end{figure*}

Figures~\ref{fig:scat-gender-age} and~\ref{fig:scat-accent-emotion} plot all systems on a shared axis, where the 95\% confidence intervals are smaller than the markers.
Figure~\ref{fig:per-system} shows the same values separately for each system.
Each panel uses its own y-axis so that the intervals are visible.
 
\section{Prompt Templates}
\label{app:prompts}
 
Each single-attribute value is realized through five templates that hold semantic content constant while varying syntactic form: canonical, possessive, participial, role attribution, and attribute-quality descriptor. Table~\ref{tab:templates-single} illustrates with two values (the male gender category and the child age category); every other single-attribute value follows the same frames with the attribute keyword swapped. Intersectional (gender$\times$age) cells use a separate 5-frame set with two word orders per frame, giving 10 templates per cell; Table~\ref{tab:templates-intersect} shows the male$\times$child cell. Templates strictly avoid prosodic modifiers (e.g., ``loudly'') to prevent confounding the intended attribute with orthogonal acoustic variations.
 
\begin{table*}[t]
\centering
\small
\renewcommand{\arraystretch}{1.25}
\setlength{\tabcolsep}{14pt}
\begin{tabular}{@{}cll@{}}
\toprule
\textbf{\#} & \textbf{Male (gender)} & \textbf{Child (age)} \\
\midrule
1 & Speak in a male voice.                            & Speak in a child's voice. \\
2 & Speak with the voice of a man.                    & Speak with the voice of a young child. \\
3 & Speak so you sound masculine.                     & Speak so you sound like a young child. \\
4 & Speak as a male speaker.                          & Speak as a child around eight years old. \\
5 & Speak with a voice that has masculine qualities.  & Speak with a voice that has childlike qualities. \\
\bottomrule
\end{tabular}
\caption{\textbf{Single-attribute prompt templates}, illustrated for the male-gender and child-age values. Gender, age, accent, and emotion each use the same five frames with the attribute keyword swapped.}
\label{tab:templates-single}
\end{table*}
 
\begin{table*}[t]
\centering
\small
\renewcommand{\arraystretch}{1.25}
\setlength{\tabcolsep}{14pt}
\begin{tabular}{@{}cll@{}}
\toprule
\textbf{\#} & \textbf{Word order} & \textbf{Template (male$\times$child)} \\
\midrule
\multirow{2}{*}{1} & age-first    & Speak as a young boy. \\
                   & gender-first & Speak as a boy who is young. \\
\cmidrule(l){2-3}
\multirow{2}{*}{2} & age-first    & Use the voice of a child who is male. \\
                   & gender-first & Use the voice of a male who is a child. \\
\cmidrule(l){2-3}
\multirow{2}{*}{3} & age-first    & Sound like a child who is male. \\
                   & gender-first & Sound like a male who is a child. \\
\cmidrule(l){2-3}
\multirow{2}{*}{4} & age-first    & Read this as a young boy. \\
                   & gender-first & Read this as a boy who is young. \\
\cmidrule(l){2-3}
\multirow{2}{*}{5} & age-first    & Take on the voice of a child who is male. \\
                   & gender-first & Take on the voice of a male who is a child. \\
\bottomrule
\end{tabular}
\caption{\textbf{Prompt templates combining gender and age attributes}, illustrated for the male child category. All gender and age combinations use ten templates derived from five sentence frames tested under both age-first and gender-first word orders.}
\label{tab:templates-intersect}
\end{table*}

\section{Artifacts: Licenses and Intended Use}
\label{app:licenses}
 
Table~\ref{tab:lic-itts} lists the seven ITTS systems in our panel, Table~\ref{tab:lic-encoder} the eleven speaker encoders, Table~\ref{tab:lic-classifier} the four controllability classifiers, and Table~\ref{tab:lic-data} the datasets referenced anywhere in the paper. All releases were verified from the respective Hugging Face or GitHub pages at the time of writing.
 
\begin{table*}[t]
\centering
\small
\setlength{\tabcolsep}{6pt}
\begin{tabular}{llcc}
\toprule
System & Release & Params & License \\
\midrule
MOSS-VoiceGenerator      & \texttt{OpenMOSS-Team/MOSS-VoiceGenerator} & 1.7B & Apache-2.0 \\
Parler-TTS Mini v1       & \texttt{parler-tts/parler-tts-mini-v1}     & 0.9B & Apache-2.0 \\
Parler-TTS Large v1      & \texttt{parler-tts/parler-tts-large-v1}    & 2.3B & Apache-2.0 \\
Qwen3-TTS Voice Design   & \texttt{Qwen/Qwen3-TTS-12Hz-1.7B-VoiceDesign} & 1.7B & Apache-2.0 \\
VoxInstruct              & \texttt{thuhcsi/VoxInstruct}               & 1.37B & MIT \\
CapSpeech-NAR            & \texttt{OpenSound/CapSpeech-models}         & 0.9B & CC~BY-NC~4.0 \\
VoxCPM2                  & \texttt{openbmb/VoxCPM2}                    & 2B & Apache-2.0 \\
\bottomrule
\end{tabular}
\caption{ITTS systems in the panel. Parameter counts exclude separately-loaded vocoder and codec modules. CapSpeech-NAR is the only non-commercial release; all generations and derived statistics for CapSpeech-NAR inherit the CC~BY-NC~4.0 restriction.}
\label{tab:lic-itts}
\end{table*}
 
\begin{table*}[t]
\centering
\small
\setlength{\tabcolsep}{6pt}
\begin{tabular}{llc}
\toprule
Encoder (code) & Toolkit / Release & License \\
\midrule
TitaNet (TIT)              & \texttt{nvidia/speakerverification\_en\_titanet\_large}                  & CC-BY-4.0 \\
SpeechBrain-ECAPA (SBEC)   & \texttt{speechbrain/spkrec-ecapa-voxceleb}                   & Apache-2.0 \\
SpeechBrain-ResNet (SBRN)  & \texttt{speechbrain/spkrec-resnet-voxceleb}                   & Apache-2.0 \\
WeSpeaker-ResNet221 (WS221)     & WeSpeaker \texttt{voxceleb\_resnet221\_LM}     & CC-BY-4.0 \\
WeSpeaker-SimAM-ResNet34 (WSVB) &
WeSpeaker \texttt{voxblink2\_samresnet34}   & CC-BY-NC-SA-4.0 \\
ERes2Net (ER2) & \texttt{iic/speech\_eres2net\_sv\_zh-cn\_16k-common} & Apache-2.0 \\
ERes2NetV2 (ER2v2)  &  \texttt{iic/speech\_eres2netv2\_sv\_zh-cn\_16k-common}        & Apache-2.0 \\
CAM++ (CAM)     &  \texttt{iic/speech\_campplus\_sv\_zh-cn\_16k-common}      & Apache-2.0 \\
x-vector (XV)              & \texttt{speechbrain/spkrec-xvect-voxceleb} & Apache-2.0 \\
WavLM-base+SV (WLM)        & \texttt{microsoft/wavlm-base-plus-sv}  & CC-BY-SA-3.0 \\
UniSpeech-SAT-base+SV (USAT) & \texttt{microsoft/unispeech-sat-base-plus-sv} & CC-BY-SA-3.0 \\
\bottomrule
\end{tabular}
\caption{Speaker encoders used in the validation study (\S\ref{sec:encoder-val}).}
\label{tab:lic-encoder}
\end{table*}

\begin{table*}[t]
\centering
\small
\setlength{\tabcolsep}{6pt}
\begin{tabular}{llc}
\toprule
Classifier & Release & License \\
\midrule
Gender (binary sex head)    & \texttt{tiantiaf/wavlm-large-age-sex} & Apache-2.0 \\
Age (regressor)             & \texttt{tiantiaf/wavlm-large-age-sex} & Apache-2.0 \\
Emotion                      & \texttt{emotion2vec/emotion2vec\_plus\_large} & Apache-2.0\textsuperscript{$\dagger$} \\
Accent (in-house)            & Trained for this paper                 & Research-only (ours; see below) \\
\midrule
\multicolumn{3}{l}{\emph{Frozen backbones used in the in-house accent classifier}} \\
WavLM-Large                 & \texttt{microsoft/wavlm-large}        & MIT \\
Whisper-Large-V3            & \texttt{openai/whisper-large-v3}      & Apache-2.0 \\
\bottomrule
\end{tabular}
\caption{Controllability classifiers. $\dagger$: the \texttt{emotion2vec\_plus\_large} model card lists Apache-2.0; the companion \texttt{\_seed} and \texttt{\_base} releases use a FunASR-linked ``model license'' instead, so re-users should re-verify from the Hugging Face page.}
\label{tab:lic-classifier}
\end{table*}
 
\begin{table*}[t]
\centering
\small
\setlength{\tabcolsep}{6pt}
\begin{tabular}{lll}
\toprule
Dataset & Use in this paper & License \\
\midrule
Emilia                        & Real-speech reference sampling (\S\ref{sec:overall}, \S\ref{sec:conditional}) & CC~BY-NC~4.0 \\
LibriSpeech & Encoder validation & CC BY 4.0 \\
CommonVoice-17                 & Accent classifier training (all 6 regions)             & CC0~1.0 \\
VoxPopuli \texttt{en\_de}/\texttt{en\_nl} & Accent classifier training (Germanic class)      & CC0 \\
AfriSpeech-200 \texttt{ZA}    & Accent classifier training (South African class)       & CC~BY-NC-SA~4.0 \\
L2-ARCTIC Hindi               & Accent classifier training (South Asia augmentation)  & CC~BY-NC~4.0 \\
\bottomrule
\end{tabular}
\caption{Datasets referenced in the paper and their licenses. Emilia, AfriSpeech-200, and L2-ARCTIC are non-commercial; derived artifacts that embed these corpora inherit their restrictions.}
\label{tab:lic-data}
\end{table*}
 
\paragraph{Intended use of existing artifacts.}
Our use of each existing artifact is consistent with its stated research or non-commercial terms: pretrained models and corpora are consumed in a controlled evaluation pipeline that generates and analyzes speech only for the measurement goals in \S\ref{sec:overall}--\S\ref{sec:conditional}. CapSpeech-NAR, Emilia, AfriSpeech-200, and L2-ARCTIC are non-commercial, so any downstream reuse of our generated audio or derived embeddings for CapSpeech-NAR, and of our Emilia-sourced reference statistics, is restricted to non-commercial research under the same terms.
 
\paragraph{Intended use of released artifacts.}
We release the evaluation code, the prompt manifests (prompts, input texts, random seeds, and sampling parameters per condition), the generated audio, and the trained accent classifier checkpoint.
The evaluation code is released under the MIT License. The prompt manifests, the generated audio, and the accent classifier checkpoint are released for non-commercial research use only (CC BY-NC 4.0), in line with the most restrictive upstream licenses they draw on (CC BY-NC 4.0, from CapSpeech-NAR).
The audio is intended for evaluating the diversity and controllability of ITTS systems; it must not be used to impersonate real speakers or to train systems that infer a person's demographic attributes from their voice.

\section{Use of AI Assistants}
\label{app:ai-use}
We used AI assistants (Claude and Gemini) for drafting and revising text, for literature search and citation checking, for writing and debugging code, and for formatting tables and figures in \LaTeX.
We also used ChatGPT 5.6 Luna to generate some of the schematic icons in Figure~\ref{fig:overview}.
The authors checked all generated text and code, verified every reported number against the experimental outputs and every citation against its original source, and take full responsibility for the content of this paper.

\section*{Acknowledgments}
This work was supported by the Ministry of Education (MOE) of Taiwan under the project Taiwan Centers of Excellence in Artificial Intelligence, through the NTU Artificial Intelligence Center of Research Excellence (NTU AI-CoRE).

\end{document}